\documentclass[smallextended]{svjour3}
\usepackage{amsmath}
\usepackage{booktabs}
\usepackage{placeins}
\usepackage{adjustbox}
\usepackage{graphicx}
\usepackage{pdflscape}
\usepackage{float}
\restylefloat{table}
\usepackage{amssymb}
\usepackage{tabu}
\usepackage{multirow}
\usepackage{rotating}
\usepackage{subfigure}
\usepackage{pdflscape}
\usepackage{epstopdf}
\usepackage{color}
\usepackage{url}
\usepackage{blindtext}
\usepackage{multicol}
\usepackage{xcolor}

\usepackage{longtable}
\floatstyle{plaintop}
\restylefloat{table}

\usepackage{geometry}

\usepackage{algorithm}
\usepackage{algorithmic}
\usepackage{ltablex}
\usepackage{siunitx}

\smartqed  
\begin{document}

\title{A binary variant of gravitational search algorithm and its application to windfarm layout optimization problem}

\titlerunning{BNAGGSA}        

\author{Susheel Kumar Joshi         \and
        Jagdish Chand Bansal
}


\institute{Susheel Kumar Joshi  \at
             South Asian University, New Delhi, India\\
              \email{sushil4843@gmail.com}           
           \and      
                  Jagdish Chand Bansal \at
             South Asian University, New Delhi, India\\
               \email{jcbansal@sau.ac.in}  
}

\date{Received: date / Accepted: date}

\maketitle

\begin{abstract}

In the binary search space, GSA framework encounters the shortcomings of stagnation, diversity loss, premature convergence and high time complexity. To address these issues, a novel binary variant of GSA called `A novel neighbourhood archives embedded gravitational constant in GSA for binary search space (BNAGGSA)' is proposed in this paper. In BNAGGSA, the novel fitness-distance based social interaction strategy produces a self-adaptive step size mechanism through which the agent moves towards the optimal direction with the optimal step size, as per its current search requirement. The performance of the proposed algorithm is compared with the two binary variants of GSA over $23$ well-known benchmark test problems. The experimental results and statistical analyses prove the supremacy of BNAGGSA over the compared algorithms. Furthermore, to check the applicability of the proposed algorithm in solving real-world applications, a windfarm layout optimization problem is considered. Two case studies with two different wind data sets of two different wind sites is considered for experiments.


\end{abstract}

\section{Introduction}

In recent years, meta-heuristic algorithms has been emerged extensively for solving complex and rigid optimization problems of both continuous and binary search space.
Transfer function is one of the basic prerequisite for converting the mechanism of an algorithm from continuous search space to binary one. Under this conversion, the search functionalities of an algorithm remain preserve. Therefore the shortcomings of an algorithm in continuous search space effect its performance in binary search space, very badly. Gravitational search algorithm (GSA) \cite{ras} is a very popular and robust optimizer for both continuous and binary search space. Rashedi et al. \cite{rashedi2010bgsa} introduced the first version of binary gravitational search algorithm (BGSA) by using the V shaped transfer function. Due to the flaws of its continuous counterpart, BGSA also suffers from the issues of stagnation occurence and the slow convergence rate. To overcome these issues and develop more efficient and robust BGSA variants, several studies have been conducted so far. In this context, Mirjalili et. al. \cite{mirjalili2014binary} proposed a self adaptive parameter scheme for a hybrid binary model associated with particle swarm optimization (PSO) and GSA. Yuan et. al. \cite{yuan2014new} proposed a local mutation strategy in BGSA for unit commitment (UC) problem. In \cite{rashedi2014feature}, a novel transfer function is introduced to improve the efficiency of BGSA for feature subset selection. In similar study, Chakraborti et. al. \cite{chakraborti2014novel} proposed a self-adaptive weight strategy in BGSA for feature selection. In another study \cite{bostani2017hybrid}, a variant of BGSA is proposed by incorporating a mutation information scheme for feature selection. Khanesar et. al. \cite{khanesar2019xor} proposed a XOR attached GSA variant for binary search space. Somu et. al. \cite{somu2020ibgss} introduced a newton-raphson inspired BGSA for QoS value prediction and service ranking prediction techniques. Guha et. al. \cite{guha2020introducing} improved exploration ability of BGSA by introducing a clustering approach for initial population distribution for feature selection problem. 
Han et. al. \cite{han2020feature} proposed a recursive BGSA for two-phase feature selection method for cancer classification. Thakur et. al. \cite{thakur2021binary} solved a multi-criteria scheduling problem by proposing a quantum-inspired BGSA.
 
Windfarm layout optimization problem is one of the most popular optimization problem in the field of renewable energy. The objective of this problem is to find the optimal layout of wind turbines in a windfarm for which the total power output and windfarm efficiency both should be maximized. So far, several studies have been conducted to optimize different objectives of the windfarm through different meta-heuristic algorithms. To minimize the cost per unit power output of a windfarm, Mosetti et al. \cite{mosetti1994optimization} proposed a binary coded genetic algorithm (GA). Grady et al. \cite{grady2005placement} improved the results of the same approach by more simulations. To improve the total power output of the windfarm, Mittal \cite{mittal2010optimization} introduced the fine grid spacing in windfarm.

Emami et al. \cite{emami2010new} optimized both cost and power of the windfarm using 
weighted aggregate method of multi-objective optimization. Marmidis et al. \cite{marmidis2008optimal} introduced Monte Carlo approximation approach in the single objective platform to study a simple wind condition. Pookpunt et. al.\cite{pookpunt2013optimal} used the binary variant of particle swarm optimization to solve the windfarm layout optimization problem. To solve the same problem, \cite{biswas2017optimal} proposed a variant of differential evolution using linear population size reduction approach. Ant colony optimization (ACO) algorithm \cite{erouglu2012design} is also used to solve the windfarm layout optimization problem. 
Literatures \cite{mosetti1994optimization} \cite{erouglu2012design} studied a basic single objection framework in which the cost function of a windfarm is optimized subject to the number of turbines. In \cite{chen2013wind}, the authors used the Greedy algorithm to optimize the layout. In a similar study, \cite{chen2016wind} used the same greedy approach for multiple hub heights of wind turbines. In \cite{gonzalez2010optimization}, evolutionary algorithm is used to optimize the positions of wind turbines for a windfarm to obtain the minimum cost model. To optimize the wind farm layout and wind turbine size, \cite{dupont2016advanced} proposed an advanced modeling system using a multi-level extended pattern search algorithm. In another study, Bansal et al. \cite{bansal2017wind} used the biogeography based optimization (BBO) for solving the windfarm layout problem. 
Biswas et. al. \cite{biswas2017optimization} used differential evolution to perform the case studies for maximizing the windfarm efficiency with different rotor diameters and hub heights. To optimize the windfarm layout, a lot of studies has been proposed by the researchers in the multi-objective framework also. In this context, Chen et al. \cite{chen2015multi} used Multi-objective genetic algorithm (MOGA) by utilizing one Pareto solution in the optimization process. In another study, \cite{feng2016multi} optimized the windfarm layout and the number of turbines both simultaneously by using multi-objective random search algorithm.
 
In a recent study \cite{biswas2018decomposition}, authors proposed an approach to solve the windfarm layout optimization problem using an advanced multi-objective evolutionary algorithm based on decomposition (MOEA/D) \cite{zhang2007moea,zhang2009performance}. In this approach, a multi-objective optimization problem is decomposed into several scalar optimization subproblems which further optimized simultaneously. Due to an ability to optimize each subproblem using information from its neighboring subproblems, MOEA/D have less computational complexity than multi-objective genetic local search (MOGLS) and nondominated sorting genetic algorithm-II (NSGA-II). In \cite{biswas2018decomposition}, authors used MOEA/D to solve windfarm layout optimization problem for maximizing the two objectives, power output and windfarm efficiency simultaneously. With this set up, several Pareto optimal solutions can be obtained by a single run of MOEA/D. The wind condition in this study considers variable wind speed and direction for two different sites \cite{biswas2018decomposition}.

This paper proposes a novel BGSA variant which is tested over different search domains of uni-modal, multi-modal and multi-modal with fix dimensions landscapes. Furthermore, the multi-objective windfarm layout optimization problem described in \cite{biswas2018decomposition} is effectively solved by the proposed variant under the priori approach. The contribution of this paper can be summarized as follows:
\begin{itemize}
\item To improve the search objectives, a novel social interaction scheme is proposed. For this, two neighbourhood archives are introduced for each candidate solution regarding its current position (F archive) and its distance (D archive) from the most promising regions of the landscape. These archives provide the most appropriate neighbours to each agent in terms of its current search requirements.         
\item A parameter-free, fitness-distance ratio based gravitational constant is proposed for a better trade-off between exploration and exploitation of the algorithm.  
\item The binary variant integrated with the above mentioned novelties under the priori approach outperforms the posteriori approach of MOEA/D for the considered multi-objective windfarm layout problem.  
\end{itemize}

Rest of the paper is organized as follows. Section \ref{sec:gsa} briefly describes the frameworks of GSA. In Section \ref{sec:proposed}, a detailed introduction of the proposed is given for continuous search space. Section \ref{sec:binary_version} describes the frameworks of proposed BNAGGSA and evaluates its performance over binary search space.  The considered windfarm layout problem is described in Section \ref{sec:windfarm}. In section \ref{Applicab}, proposed BNAGGSA algorithm is applied to solve the considered windfarm layout optimization problem along with the experimental settings and simulation results. Finally, Section \ref{sec:con} concludes the paper.

\section{Basic Gravitational Search Algorithm}\label{sec:gsa}
\vspace*{1.5px}
In the recent years, Gravitational Search Algorithm (GSA) \cite{ras} has been widely used meta-heuristic in different domains of science, engineering and research. It cleverly mimics the law of gravity and mass interaction and produces a robust search mechanism as follows:

Initially, a swarm of $N$ particles is randomly generated within the predefined search space $\mathbb{S}$ having $n$ dimensions. Each particle ($X_i, i=1,2,...,N$) of the generated swarm is defined as:   

\begin{equation}\label{eq1}
  X_{i}=(x_{i}^{1},.....,x_{i}^{d},.....,x_{i}^{n}), ~~ \forall ~i=1,2,.....,N
\end{equation}

After obtaining the fitness values ($fit_i(t), i=1,2...,N$) at a specific iteration $t$, the mass $M_i(t)$ of each $X_i$ is calculated as: 

\begin{equation}
q_i(t)=\frac{fit_i(t)-worst(t)}{best(t)-worst(t)}
\end{equation}
\begin{equation}
M_i(t)=\frac{q_i(t)}{\sum_{j=1}^{N}q_j(t)}, ~~\forall ~i=1,2,.....,N
\end{equation}
Here best(t) and Worst(t) are the best and worst fitness of the current swarm,  respectively.

The acceleration $a_i^{d}(t)$ of $X_i$ in $d^{th}$ dimension is defined as: 
\begin{equation}
a_i^{d}(t)=\frac{F_i^{d}(t)}{M_i(t)}
\end{equation}

Here, $F_i^{d}(t)$ is the total force acting on $X_i$ by a set named $K_{best}$ archive having $K$ biggest masses associated with $K$ best fit particles of the current swarm. 
$F_i^{d}(t)$ is calculated as:

\begin{equation}
F_{i}^{d}(t) = \sum _{j \in K_{best}, j \neq i}rand_{j} \times F_{ij}^{d}(t)
\end{equation}
Here, the $K_{best}$ set is solely responsible for the social interaction in GSA framework. Its size reduces from $K$ to $1$ iteratively. $rand_{j}$ is a uniform random number from the interval $[0, 1]$. At a specific iteration $t$, $F_{ij}^{d}(t)$ is the force between $i^{th}$ and $j^{th}$ particles in the ${d}^{th}$ dimension and calculated as:

\begin{equation}\label{eq:force}
F_{ij}^{d}(t)=G(t)\frac{M_i(t)M_j(t)}{R_{ij}+\epsilon } (x_i^{d}(t)-x_j^{d}(t))
\end{equation}

Here, $G(t)$, a scaling factor called gravitational constant is an exponential deceasing function defined as: 
\begin{equation}\label{eq:gc}
G(t)=G_{0}e^{-\alpha \frac{t}{T}}
\end{equation}
$G_{0}$ and $\alpha$ are fixed values set to $100$ and $20$, respectively. $T$ is the maximum number of iterations. $R_{ij}(t)$ is distance between particles $X_i$ and $X_j$. $\epsilon$ is a very small positive value. Now, $a_i^{d}(t)$, the acceleration $a_i^{d}(t)$ of $X_i$ in $d^{th}$ dimension is defined as:  
\begin{equation}\label{eq:acc}
a_i^{d}(t)=\sum_{j\in K_{best},j\neq i}rand_{j}G(t)\frac{M_j(t)}{R_{ij}+\epsilon } (x_i^{d}(t)-x_j^{d}(t)),
\end{equation}
$d = 1, 2, . . . , n$ and $i = 1, 2, . . . , N$.\\
The acceleration $a_i^{d}(t)$ helps to update the position of $X_i$ through the following bootstrapping equations:  
\begin{equation}\label{eq:vel}
v_{i}^{d}(t+1)=rand_i\times{v_{i}^{d}(t)}+a_{i}^{d}(t)
\end{equation}
\begin{equation}\label{eq:pos}
x_{i}^{d}(t+1)=x_{i}^{d}(t)+v_{i}^{d}(t+1)
\end{equation}
Here, $rand_{i}$ is uniformly selected random number from the interval $\left[0,1\right]$. $v_{i}^{d}(t)$ and $x_{i}^{d}(t)$ are the velocity and position of $X_i$ in $d^{th}$ dimension,  respectively.

\section{ The proposed GSA variant for continuous search space}\label{sec:proposed}
In the basic GSA framework, $K_{best}$ set or archive is the neighbourhood structure through which the particles interact. This structure provides the same $K$ sub-optimal regions of the fitness landscape to each particle for a significant search towards optimality. However, this structure ignores the other valuable informations associated with the current position of the particle. This demerit of $K_{best}$ archive makes GSA search mechanism vulnerable in terms of diversity control. The gravitational constant $G$, on the other hand, is another entity of GSA model which entirely responsible for scaling the step sizes of its particles. Although $G$ works significantly well due to its exponential decreasing property, it also does not consider the particle's individual requirements and scales the next moves of all the particles in the same fashion. This shortcoming produces inappropriate step sizes which further causes either stagnation or sometimes bypassing the true optima. 

To overcome these mentioned issues of $K_{best}$ archive and $G$, authors have proposed a novel variant of GSA, namely `A novel neighbourhood archives embedded gravitational constant in GSA (NAGGSA)' \cite{joshi2021novel}. In NAGGSA, first, two novel neighbourhood archives are proposed for each particle to consider its other informative characteristics as per its current position. Secondly, a novel scaling mechanism is proposed which scales the next move of a particle according to its current search requirements. To check its performance in the continuous search space, authors have done all the necessary experiments and analyses through which it proves its efficiency in the continuous search space. The detailed description about the mechanism of proposed NAGGSA can be found in \cite{joshi2021novel}. 

\section{GSA and NAGGSA in the binary search space}\label{sec:binary_version}
 In the binary search space, particle updates itself through switching between $0$ and $1$ values. In this study, the mentioned switching is done by a well-known sigmoid function which defines a least probability of particle's movement for small $(V_i)$ while for a large $(V_i)$, it provides a high probability for particle's movement \cite{rashedi2010bgsa}. The sigmoid function is defined as:

\begin{equation}
S(V_i)=|tanh(V_i)|
\end{equation}

After calculating $S(V_i)$, particle's movement is decided by the following rules:  
\begin{equation}
\begin{split}
if~rand < S(V_i) \Rightarrow X_i^d(t+1)=complement( X_i^d(t))\\
else~X_i^d(t+1)= X_i^d(t)
\end{split}
\end{equation}

\begin{figure}
 \centering
  \subfigure[]{
 \includegraphics[height=4.5cm, width=6.9cm]{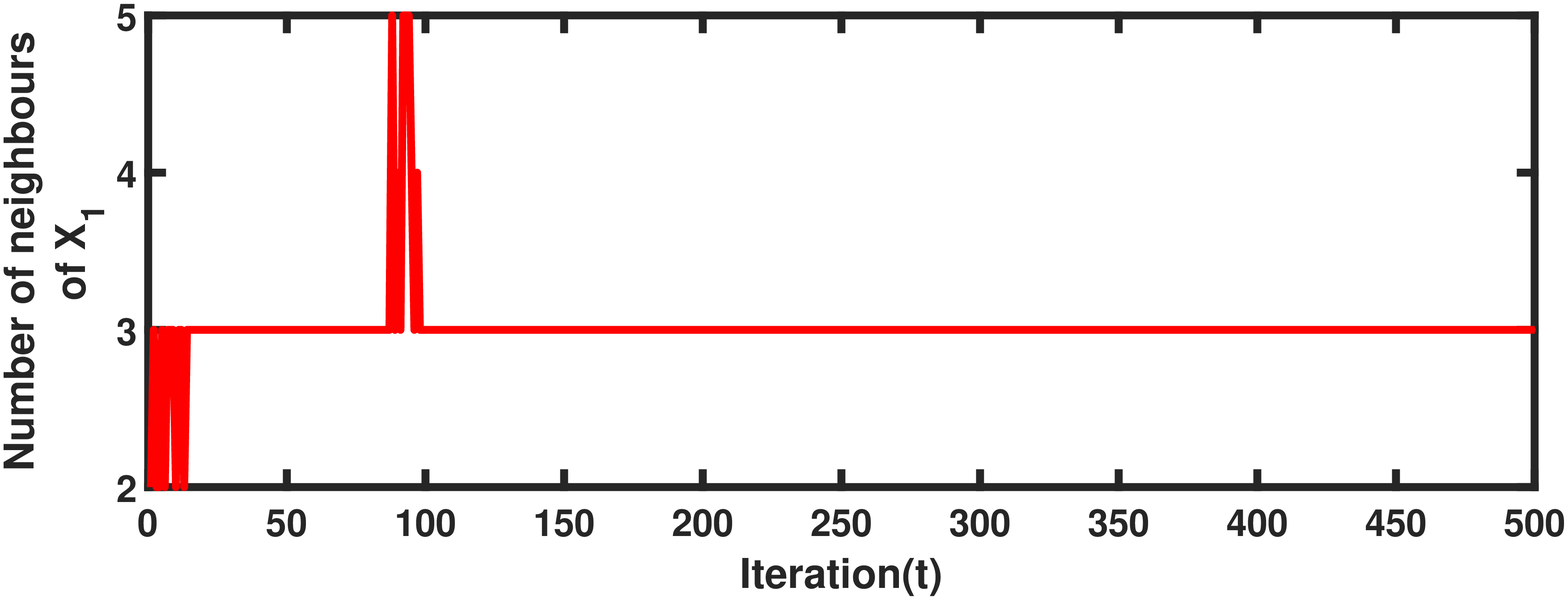}}
  \subfigure[]{
    \includegraphics[height=4.5cm, width=6.9cm]{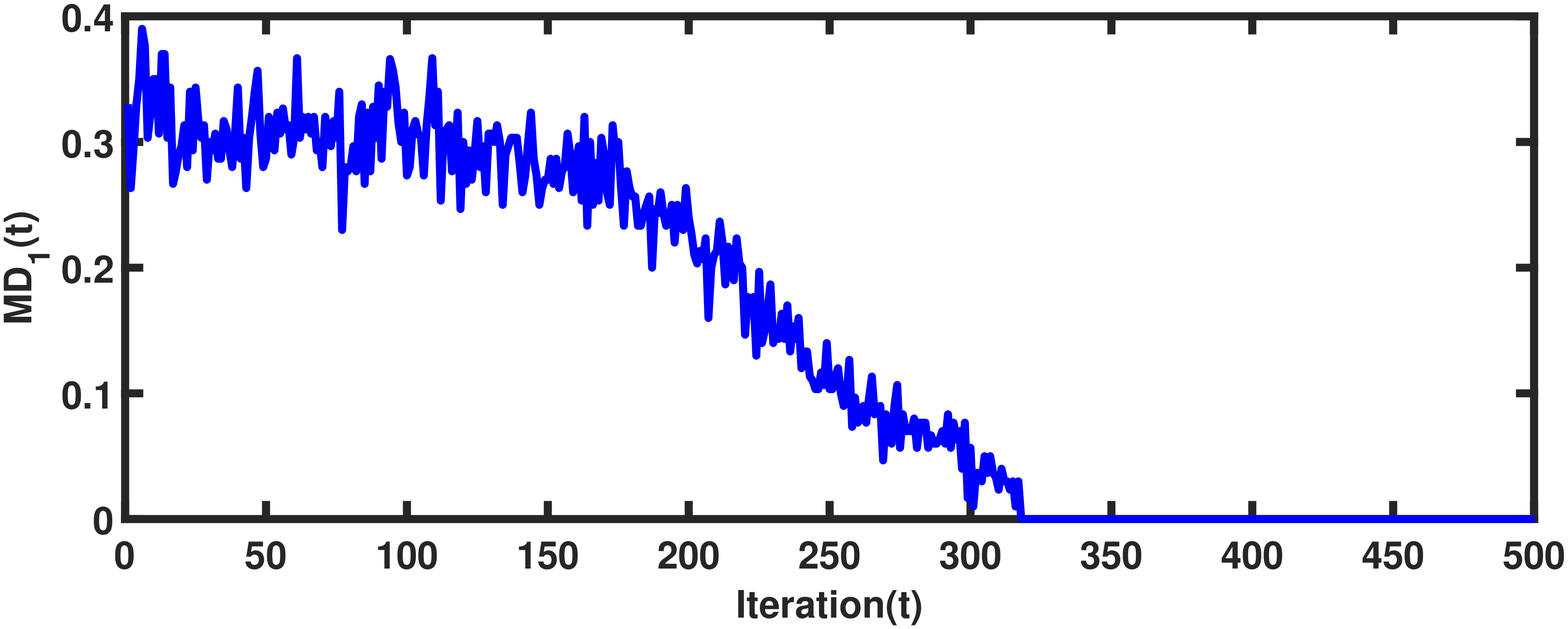}}
 \subfigure[]{
    \includegraphics[height=4.5cm, width=6.9cm]{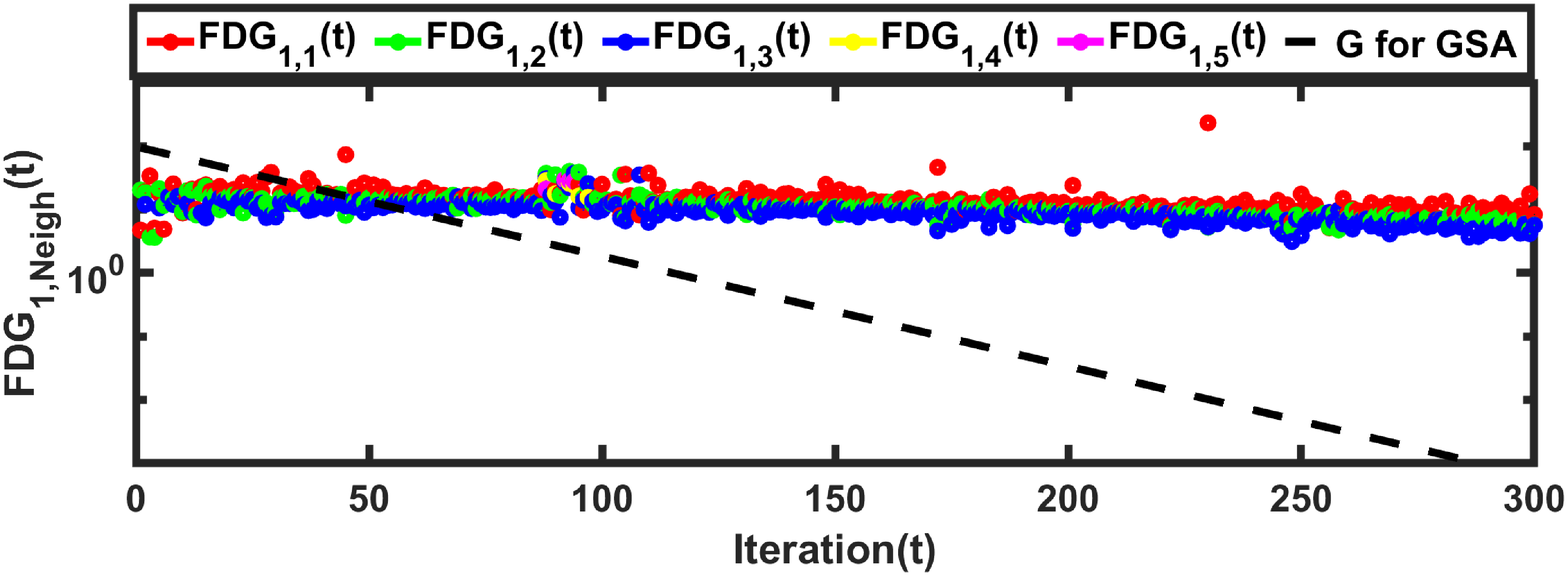}}
 \subfigure[]{
    \includegraphics[height=4.5cm, width=6.9cm]{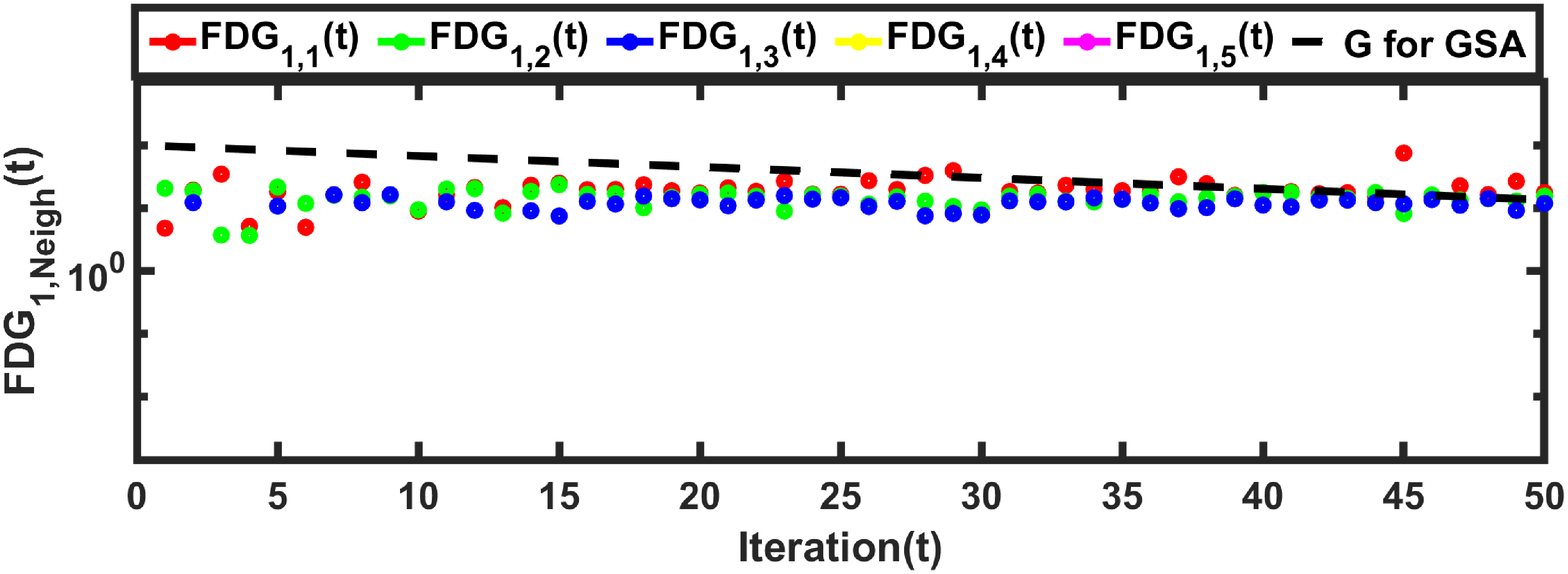}}
    \caption{Graphical analyses of first particle $X_1$ through its  different attributes  over $f_1$ (uni-modal function $f_1$) } 
 \label{fig:f1}
 \end{figure}

\begin{figure}
\centering
 \subfigure[]{
  \includegraphics[height=4.5cm, width=6.9cm]{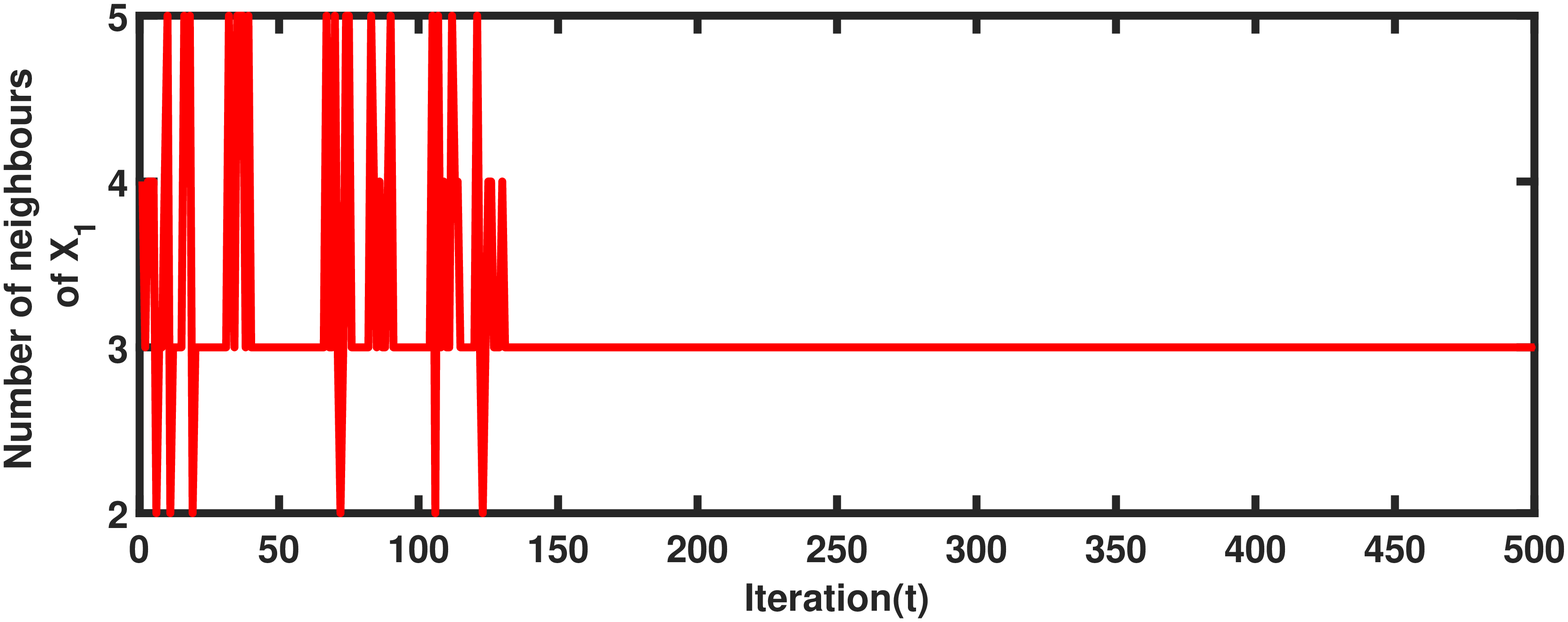}}
 \subfigure[]{
    \includegraphics[height=4.5cm, width=6.9cm]{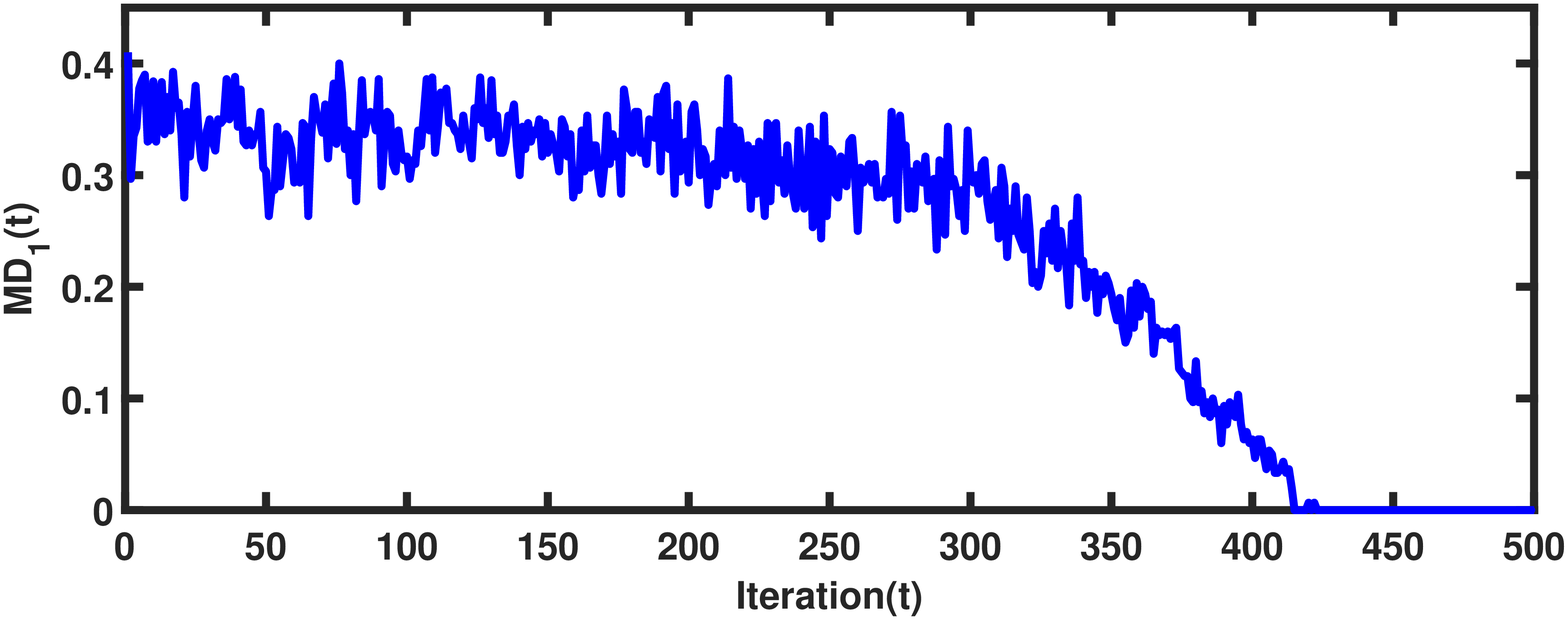}}
 \subfigure[]{
    \includegraphics[height=4.5cm, width=6.9cm]{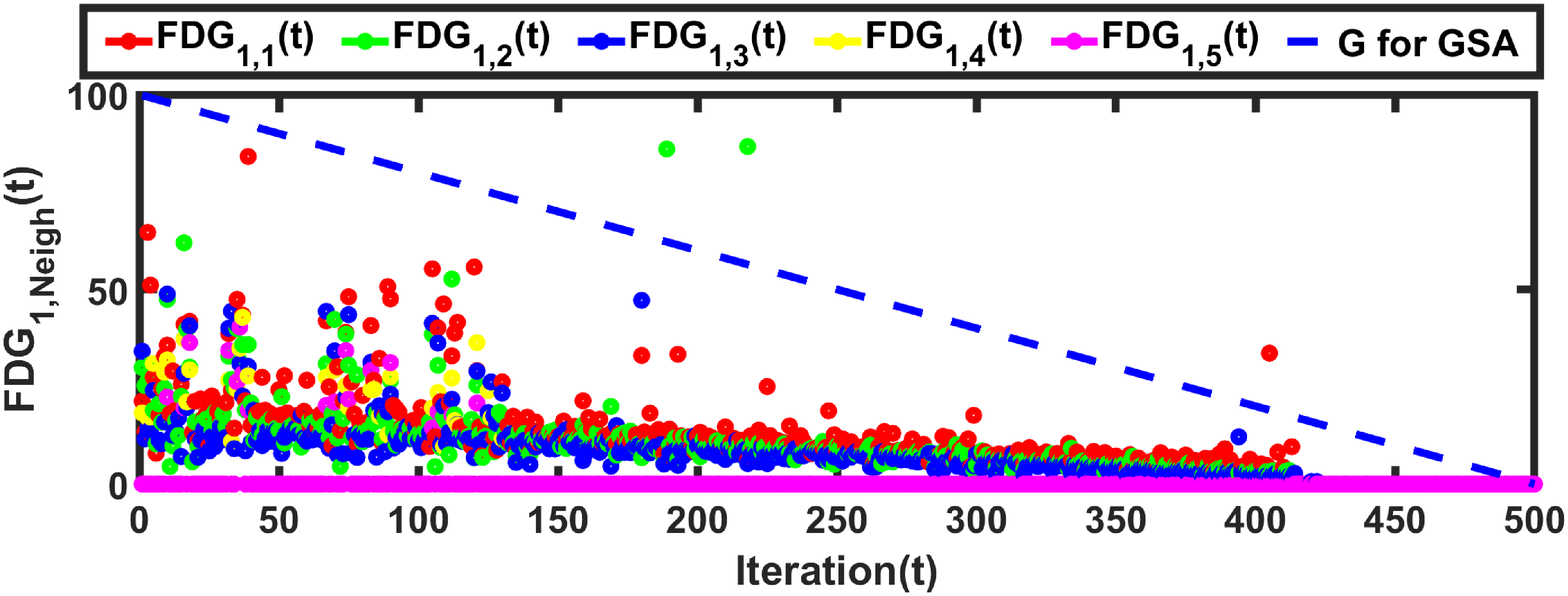}}
 \subfigure[]{
    \includegraphics[height=4.5cm, width=6.9cm]{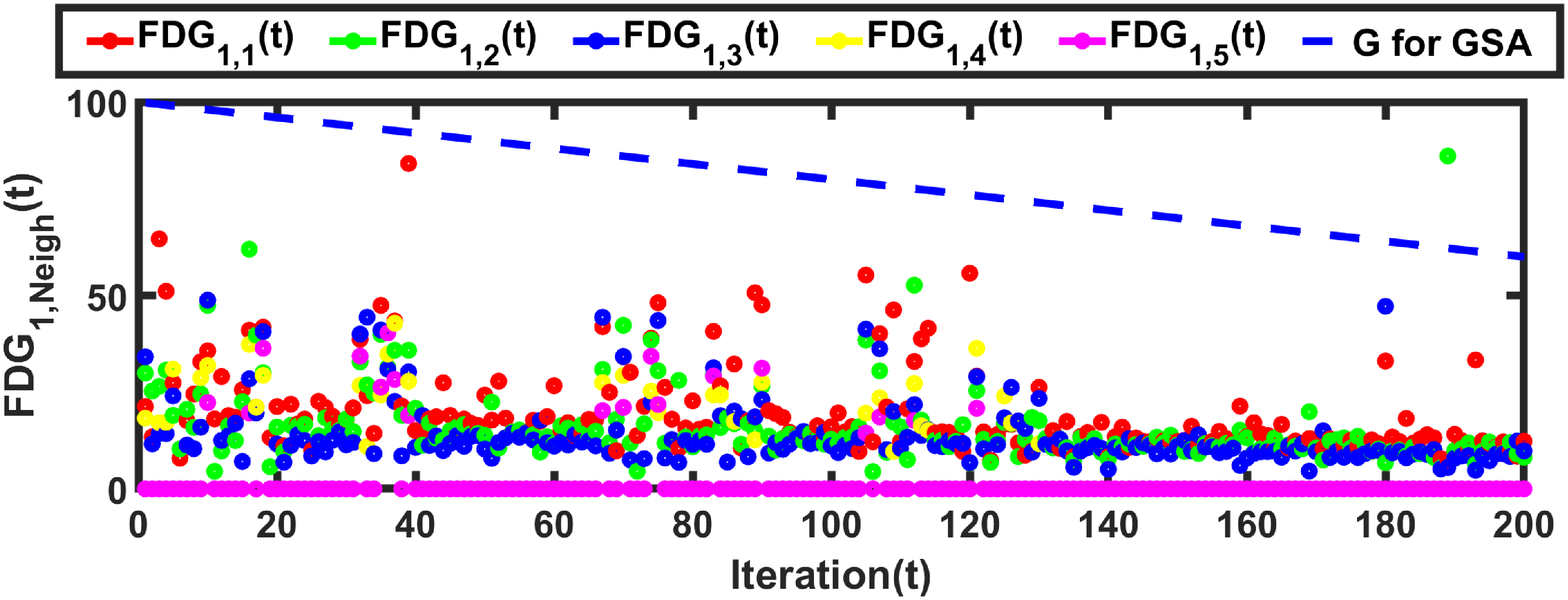}}
  \caption{ Graphical analyses of first particle $X_1$ through its  different attributes  over $f_8$ (Multimodal function) } 
 \label{fig:f8}
  \end{figure}

\begin{figure}
 \centering 
  \subfigure[]{
   \includegraphics[height=4.5cm, width=6.9cm]{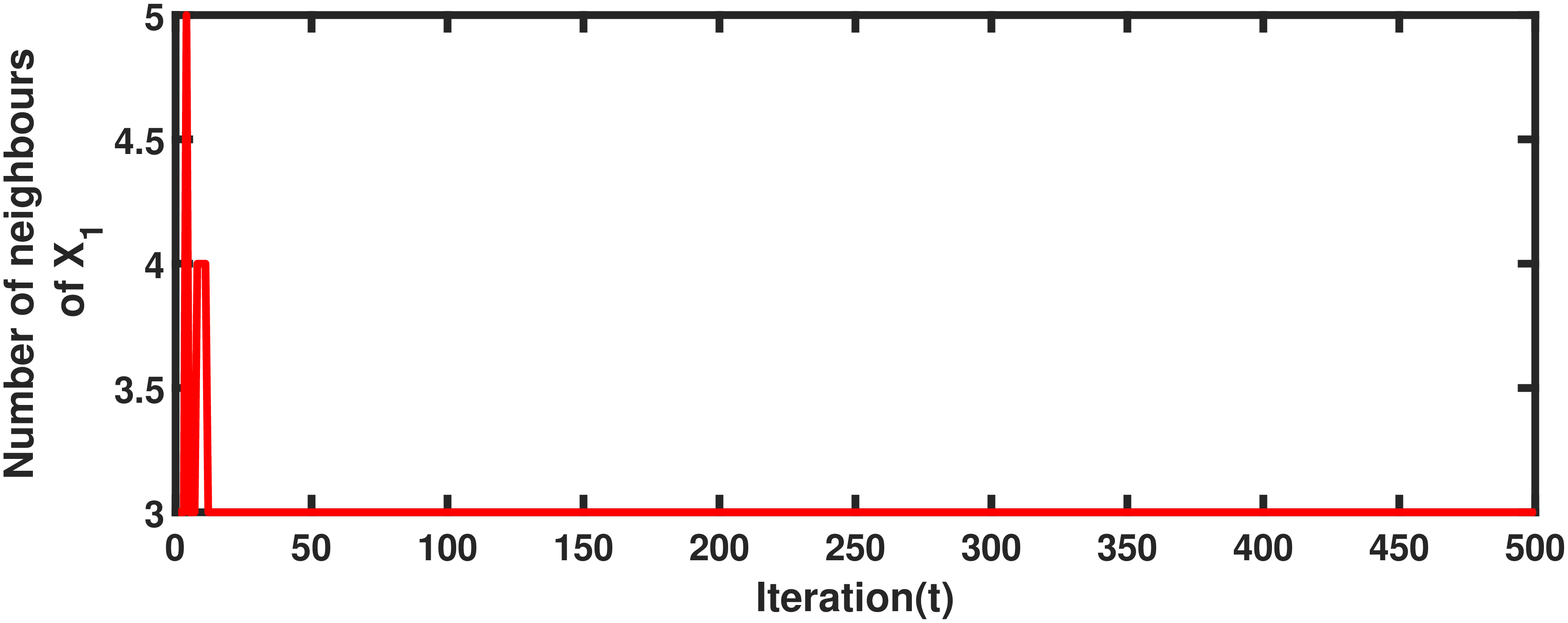}}
 \subfigure[]{
    \includegraphics[height=4.5cm, width=6.9cm]{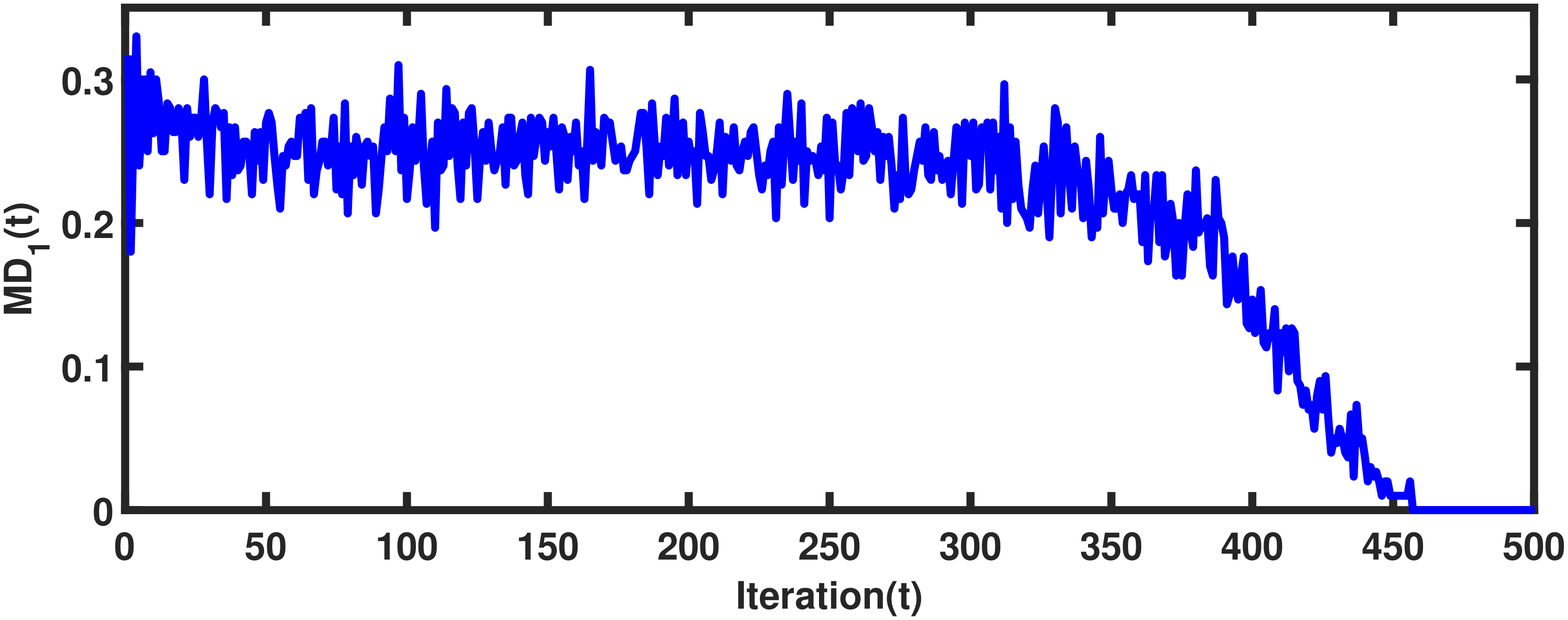}}
\subfigure[]{
    \includegraphics[height=4.5cm, width=6.9cm]{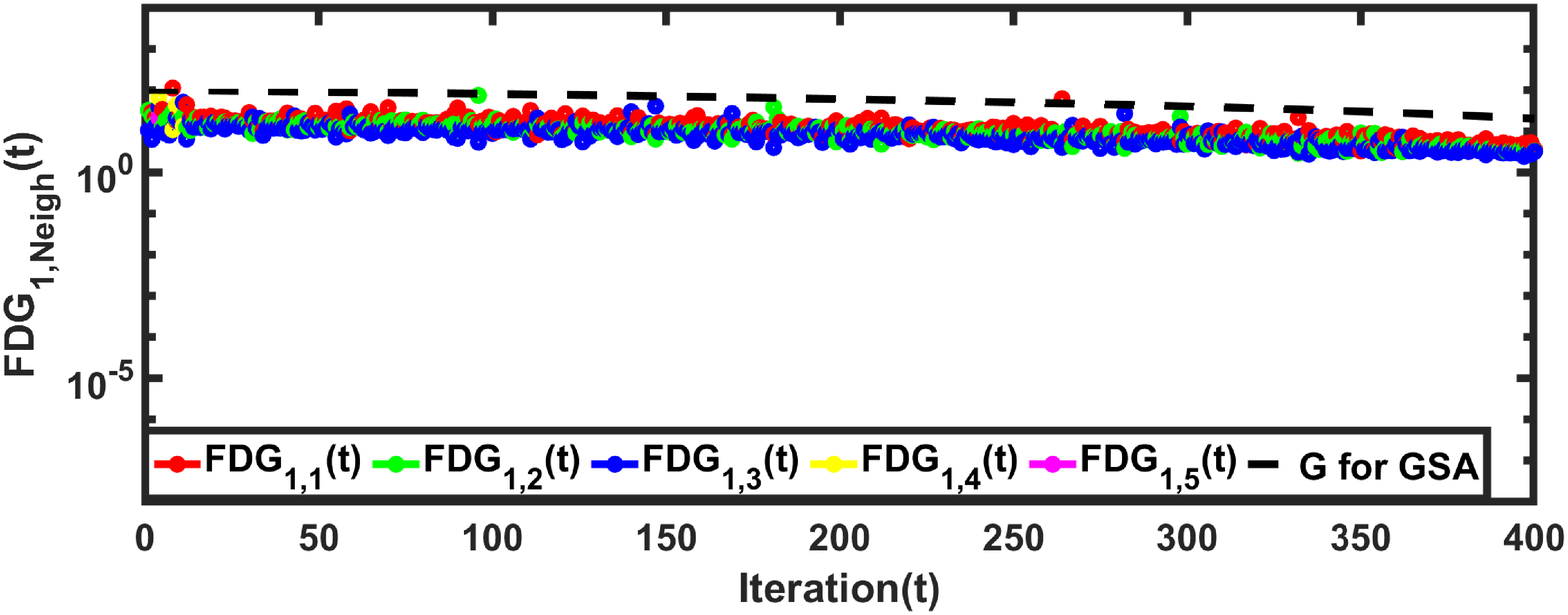}}
\subfigure[]{
    \includegraphics[height=4.5cm, width=6.9cm]{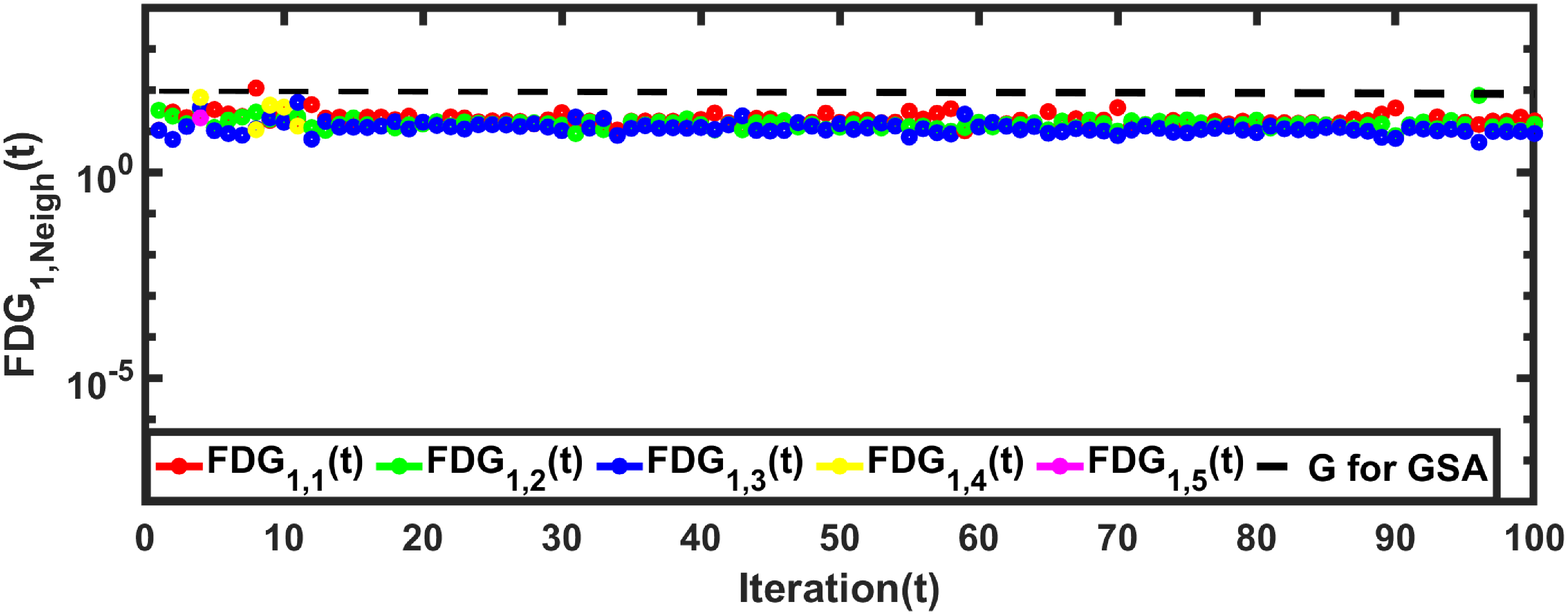}}
  \caption{Graphical analyses of first particle $X_1$ through its  different attributes  over $f_{23}$ (Multimodal test problem with fixed dimension)} 
 \label{fig:f23}
  \end{figure}

Here, rand is the normally distributed random number between $0$ and $1$. In the binary search space, the distance between two particles (considered as strings) is measured by the hamming distance which count the minimum number of substitutions required to coincide both the particles (both strings). Additionally, for a better convergence, the particles velocity is bounded in $[-6,6]$. Both GSA and NAGGSA follow the above procedure to convert their self into BGSA and BNAGGSA, respectively. To understand the overall search mechanism of the proposed BNAGGSA graphically, the first particle $X_1$ of the swarm is considered through its different attributes. Figures \ref{fig:f1}, \ref{fig:f8} and \ref{fig:f23} illustrate these graphical analyses for $f_{1}$ (unimodal function), $f_{8}$ (multimodal function) and $f_{23}$ (multimodal function with fixed dimension) (refer section \ref{sec:testproblem}). Each mentioned figures have four subfigures from (a) to (d). Subfigure (a) presents the number of neighbours $X_1$ gets by the proposed archives (either $F$ or $D$) all over the search. Each subfigure (a) also indicates that a particle can have $2$ to $5$ neighbours in each iteration. Subfigure (b) and Subgraph (c) illustrate the behaviour of the mean distance 
and the proposed $FDG_{1,Neigh}(t)$ of $X_1$ from its assigned neighbours, respectively.
For a better comparison, gravitational constant $G$ of BGSA (mentioned in section \ref{Section:setting}) is also considered in the Subfigure (c). Subfigure (d) is just a magnified version of subfigure (c).


\subsection{Result and Discussion}\label{sec:testproblem}
\subsubsection{Test problems under consideration}\label{routine}
In order to validate the performance of the proposed binary variant BNAGGSA, a set of well known continuous benchmark problems is considered and listed in Table \ref{table:continuous}. In the Table \ref{table:continuous}, $C$ represents the topological characteristics of the test problem through which the test problems are categorized into three groups: unimodal problem ($U$), multimodal problem ($M$) and multimodal problem with fix dimension ($MFD$). $m$ is the dimension of the test problem, $f_{min}$ is the optimal value of the test problem and $S$ is the subset of  $\mathbb{R}^m$. As the continuous benchmark problems (refer Table \ref{table:continuous}) are considered to evaluate the performance of the proposed binary variant. It is required to have a strategy through which the binary representation can convert to real values. For this, first, a swarm having $N$ particles is generated in such a way that its each particle can represent as a vector of $m$ bit strings ($m$ is the dimension of the test problems of Table \ref{table:continuous}). Each bit have length $L$ which is set to  $15$ here. In this study, the following simple strategy is used to convert each bit string into an integer:
 
\begin{equation}
X_{integer}=\sum_{i=0}^{L}(X_i\times2^i)
\end{equation} 
 
Here, $X_i$ is the $i^{th}$ bit of $X$ in its binary format. The obtained integer is further converted into its corresponding real value within the continuous interval $[a,b]$ as follows:

\begin{equation}
X_{Real}=a+X_{integer}\times\frac{(b-a)}{2^L}
\end{equation}

\newpage
\begin{center}
\begin{small}
\setlength{\LTleft}{-20cm plus -1fill}
\setlength{\LTright}{\LTleft}

\begin{longtable}{p{11cm}p{2.0cm}p{1.7cm}p{1.4cm}}

 \caption{Test problems under consideration}
 \label{table:continuous}
\endfirsthead
\caption* {\textbf{Table \ref{table:continuous} Continued:}}
\endhead
\toprule
 \textbf{Test problems} &   \textbf{S}  & \textbf{$f_{min}$}  &\textbf{C}  \\
 \midrule

$f_1(x)=\sum_{i=1}^{n}x_{i}^{2}$  & ${[}$-$100,100{]}^m$ & 0 & $U$\\[1ex]

 $f_2(x)=\sum_{i=1}^{n}|x^{2}_{i}|$+ $\prod_{i=1}^{n}|x_{i}|$  & ${[}$-$10,10{]}^m$ & 0 & $U$\\[1ex]
 
 $f_3(x)=\sum_{i=1}^{n}\left(\sum_{j-1}^{i}x_{j}\right)^{2}$  & ${[}$-$100,100{]}^m$ & 0 & $U$\\[1ex]

 $f_4\left(x\right)=\max_{i}\left\{ |x_{i}|,1\leq i\leq n\right\} $  & ${[}$-$100,100{]}^m$ & 0 & $U$\\[1ex]

 $f_5(x)=\sum_{i=1}^{n-1}\left[100\left(x_{i+1}-x_{i}^{2}\right)^{2}+\left(x_{i}-1\right)^{2}\right]$  & ${[}$-$30,30{]}^m$ & 0& $U$\\[1ex]

 $f_6(x)=\sum_{i=1}^{n}\left(\left[x_{i}+0.5\right]\right)^{2}$  & ${[}$-$100,100{]}^m$ & 0 & $U$\\[1ex]
 
 $f_7(x)=\sum_{i=1}^{n}ix_{i}^{4}+random[0,1)$  & ${[}$-$1.28,1.28{]}^m$ & 0 & $U$\\[1ex]
 
  $f_8(x)=\sum_{i=1}^{n}-x_{i}\sin\left(\sqrt{|x_{i}|}\right)$  & ${[}$-$500,500{]}^m$ & $-$418.9829$\times m$  & $M$\\ [1ex]
  
  $f_9(x)$=$\sum_{i=1}^{n}\left[x_{i}^{2}-10\cos\left(2\pi x_{i}\right)+10\right]$  & ${[}$-$5.12,5.12{]}^m$ & 0 & $M$\\[1ex]
  
  $f_{10}(x)=-20\exp(-0.2\sqrt{\frac{1}{n}\sum_{i=1}^{n}x_{i}^{2}})-\exp\left(\frac{1}{n}\sum_{i=1}^{n}cos\left(2\pi x_{i}\right)\right)+20+e$ & ${[}$-$32,32{]}^m$ & 0& $M$\\ [1ex]
  
  $f_{11}(x)=\frac{1}{4000}\sum_{i=1}^{n}x_{i}^{2}-\prod_{i=1}^{n}\cos\left(\frac{x_{i}}{\sqrt{i}}\right)+1$  & ${[}$-$600,600{]}^m$ & 0 & $M$\\[1ex]

    \bottomrule
 \pagebreak
 \toprule
\textbf{Test problems} &    \textbf{Range}  & \textbf{$f_{min}$}  &\textbf{C}  \\
 \midrule   
  
  $f_{12}(x)=\frac{\pi}{n}\left\{ 10\sin\left(\pi y_{1}\right)+\sum_{i=1}^{n-1}(y_{i}-1)^{2}\left[1+10\sin^{2}(\pi y_{i+1})\right]+(y_{n}-1)^{2}\right\} +\sum_{i=1}^{n}u(x_{i},10,100,4)$\\
  $y_{i}=1+\frac{x_{i}+1}{4}u(x_{i},a,k,m)=\begin{cases}
  k(x_{i}-a)^{m} & x_{i}>a\\
  0-a & <x_{i}<a\\
  k(-x_{i}-a)^{m} & x_{i}<-a
  \end{cases}$  & ${[}$-$50,50{]}^m$ & 0 & $M$\\[1ex]

  $f_{13}(x)=0.1\left\{ \sin^{2}(3\pi x_{1})+\sum_{i=1}^{n}\left(x_{i}-1\right)^{2}\left[1+\sin^{2}(3\pi x_{i}+1)\right]+(x_{n}-1)^{2}\left[1+\sin^{2}(2\pi x_{n})\right]\right\} +\sum_{i=1}^{n}u(x_{i},5,100,4)$ &  ${[}$-$50,50{]}^m$ & 0 & $M$\\[1ex]

  $f_{14}(x)=\left(\frac{1}{500}+\sum_{j=1}^{25}\frac{1}{j+\sum_{i=1}^{2}\left(x_{i}-a_{ij}\right)^{6}}\right)^{-1}$ &  $\left[-65,65\right]^2$ & 0.998 & $MFD$\\[1ex]

    $f_{15}(x)=\sum_{i=1}^{11}\left[a_{i}-\frac{x_{1}\left(b_{i}^{2}+b_{i}x_{2}\right)}{b_{i}^{2}+b_{i}x_{3}+x_{4}}\right]^{2}$ &  $\left[-5,5\right]^4$ & 0.00030 & $MFD$\\[1ex]
    
    $f_{16}(x)=4x_{1}^{2}-2.1x_{1}^{4}+\frac{1}{3}x_{1}^{6}+x_{1}x_{2}-4x_{2}^{2}+4x_{2}^{4}$ & $\left[-5,5\right]^2$ & $-$1.0316 & $MFD$\\[1ex]
    
    $f_{17}\left(x\right)=\left(x_{2}-\frac{5.1}{4\pi^{2}}x_{1}^{2}+\frac{5}{\pi}x_{1}-6\right)^{2}+10\left(1-\frac{1}{8\pi}\right)\cos x_{1}+10$ &  $\left[-5,5\right]^2$ & 0.398 & $MFD$\\[1ex]
    
    $f_{18}(x)=\left[1+\left(x_{1}+x_{2}+1\right)^{2}\left(19-14x_{1}+3x_{1}^{2}-14x_{2}+6x_{1}x_{2}+3x_{2}^{2}\right)\right]$\\ $\left[30+\left(2x_{1}-3x_{2}\right)^{2}\left(18-32x_{1}+12x_{1}^{2}+48x_{2}-36x_{1}x_{2}+27x_{2}^{2}\right)\right]$ &  $\left[-2,2\right]^2$ & 3 & $MFD$\\[1ex]

    $f_{19}(x)=-\sum_{i=1}^{4}c_{i}\exp\left(-\sum_{j=1}^{3}a_{ij}\left(x_{j}-p_{ij}\right)^{2}\right)$ &  $\left[1,3\right]^3$ & $-$3.86 & $MFD$\\[1ex]
    
    $f_{20}(x)=-\sum_{i=1}^{4}c_{i}\exp\left(-\sum_{j=1}^{6}a_{ij}\left(x_{j}-p_{ij}\right)^{2}\right)$ &  $\left[0,1\right]^6$ & $-$3.32 & $MFD$\\[1ex]
    
    $f_{21}(x)=-\sum_{i=1}^{5}\left[(X-a_{i})\left(X-a_{i}\right)^{T}+c_{i}\right]^{-1}$ & $\left[0,10\right]^4$ & $-$10.1532 & $MFD$\\[1ex]
    
    $f_{22}(x)=-\sum_{i=1}^{7}\left[(X-a_{i})\left(X-a_{i}\right)^{T}+c_{i}\right]^{-1}$ &  $\left[0,10\right]^4$ & $-$10.4028 & $MFD$\\[1ex]
    
    $f_{23}(x)=-\sum_{i=1}^{10}\left[(X-a_{i})\left(X-a_{i}\right)^{T}+c_{i}\right]^{-1}$ &  $\left[0.10\right]^4$ & $-$10.5363 & $MFD$\\[1ex]

\bottomrule
\end{longtable}

 \end{small}
 \end{center}


\subsection{Experimental setting}
\label{Section:setting}
In order to validate the effectiveness and robustness of the proposed algorithm, BNAGGSA is compared with BGSA \cite{rashedi2010bgsa}, a binary version of 
FVGGSA \cite{Bansal2018} and a binary version of PTGSA \cite{joshi2019parameter} with the following experiment setting:

\begin{itemize}
\item  The number of simulations/run =$30$,
\item  Swarm size=$50$, 
\item  For the binary search space, the dimension of $f_1$ to $f_{13}$ in Table \ref{table:continuous} is $m$ $\times$ $L$. Here, $m$ and $L$ are set to $5$ and $15$ respectively. Therefore, all the binary version of the considered uni-modal ($f_1$-$f_7$) and multi-modal ($f_8$-$f_{13}$) problems have $75~(5\times 15)$ dimensional search space. On the other hand, the binary version of multi-modal with fixed dimensional problems ($f_{14}$-$f_{23}$) have different dimensions from the range $[2\times15, 6\times 15]$. For $f_{14}$, $f_{16}$, $f_{17}$ and $f_{18}$, the dimension is $30~(2\times15)$. For $f_{15}$, $f_{21}$, $f_{22}$ and $f_{23}$, the dimension is $60~(4\times15)$. $f_{19}$ has $45~(3\times 15)$ dimensional search space while $f_{20}$ is the the maximum dimensional problem having $90~(6\times15)$ dimension.            

\item Maximum number of iteration =$500$ 

\item Parameters for the algorithms BGSA are same as GSA except the gravitational constant $G$. In  BGSA, $G$ is linear in nature defined as:
\begin{equation}
G(t)=G_0\times(1-\frac{t}{T})
\end{equation}
Here, $t$ and $T$ have their usual meaning. The parameters of the proposed variant and other considered algorithms are listed in Table \ref{table:parameters}. 

\item  The routine which converts binary representation into real, discussed in Section \ref{routine}, is used for all considered algorithms.
\end{itemize}

\subsection{Results and statistical analysis}

\begin{figure}
   \centering
 \subfigure[For unimodal test problem $f_1$]{
    \includegraphics[height=4.5cm, width=12cm]{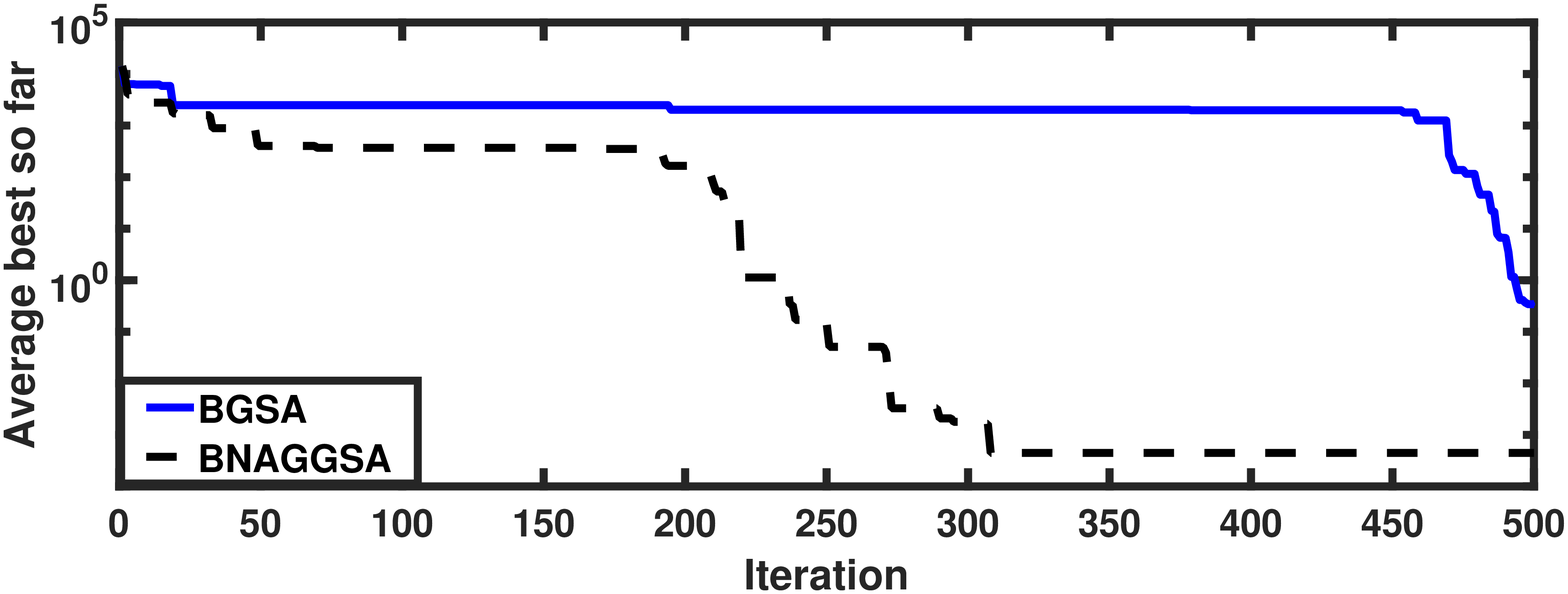}}
    \label{fig:uni_f1}
\subfigure[For multimodal test problem $f_8$]{
   \includegraphics[height=4.5cm, width=12cm]{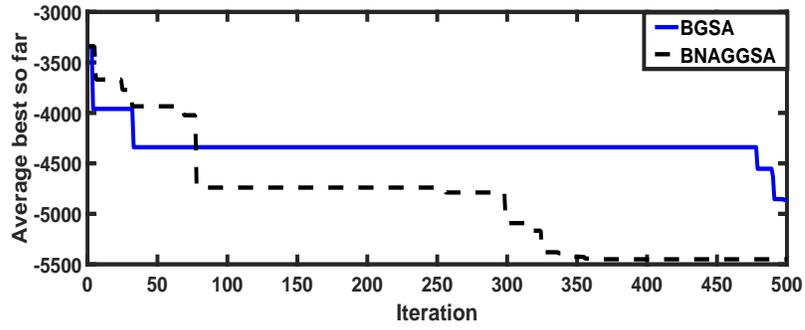}}
    \label{fig:multi_f8}
\subfigure[For multimodal problem $f_{23}$ with fixed dimension]{
    \includegraphics[height=4.5cm, width=12cm]{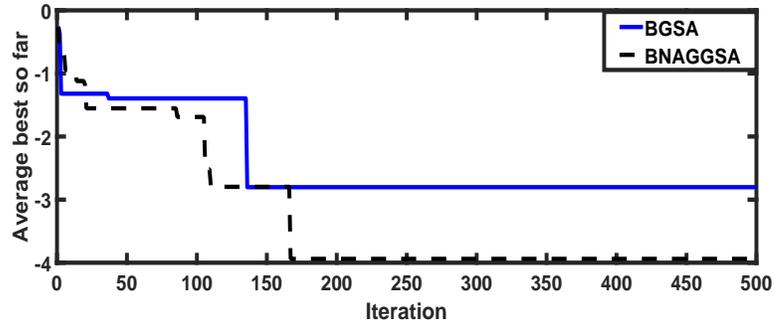}}
    \label{fig:fixed_f23} 
  \caption{Convergence graphs}   
 \label{fig:convergence}
  \end{figure}

The experimental results of best-so-far solution over $30$ independent runs under $30$ different random seeds are summarized in Table \ref{table:results}. The bold entries show the best results. Table \ref{table:results} lists the three metrics of best-so-far solution: average best-so-far solution ($ABSF$), standard deviation ($STDV$) of best solution and the best obtained solution over $30$ runs ($best$). The best-so-far solution is the solution of the last iteration of an individual run. For all unimodal test problems ($f_1$-$f_7$), the proposed BNAGGSA outperforms other BGSA variants with a large margin. It proves the fast convergence ability of the proposed variant. Out of six multimodal problems ($f_8$-$f_{13}$), all six $ABSF$ and $STDV$ values of the proposed BNAGGSA proves its supreme stagnation avoidance mechanism over others. Since the multimodal problems have multiple local optima around the global optimum in which the swarm can stagnate into the non-optimal basin. While in BNAGGSA, the proposed fitness distance ratio based gravitational constant provides a dynamic mechanism as per the search necessity which successfully enables the population to track more promising regions. Out of $10$ multimodal problems with fix dimension ($f_{14}$-$f_{23}$), BNAGGSA outperforms others in terms of $ABSF$ on $9$ problems ($f_{15}$, $f_{16}$, $f_{17}$, $f_{18}$, $f_{19}$, $f_{20}$, $f_{21}$, $f_{22}$, and $f_{23}$). Among all metrics of comparison, BNAGGSA proves its supremacy on five test problems ($f_{15}$, $f_{16}$, $f_{18}$, $f_{19}$, and $f_{22}$). Figure \ref{fig:convergence} illustrates the convergence behaviors of BNAGGSA and BGSA over unimodal ($f_1$), multimodal ($f_8$) and multimodal with fixed dimension ($f_{23}$) function. It is clear from Figure \ref{fig:convergence} that BNAGGSA outperforms BGSA in terms of exploitation ability due to its fastest convergence rate. Further, to compare the algorithms based on \emph{ABSF} over all $23$ test problems, boxplot analysis has been carried out.

\begin{center}
\begin{small}
\setlength{\LTleft}{-20cm plus -1fill}
\setlength{\LTright}{\LTleft}
\begin{longtable}{p{2cm}p{8cm}}
 \caption{Parameters of the considered algorithms}
 \label{table:parameters}
\endfirsthead
\caption* {\textbf{Table \ref{table:continuous} Continued:}}
\endhead
\toprule
 \textbf{Algorithm} &   \textbf{Parameters}  \\
 \midrule
 BGSA    & independent of $\alpha$, $G_0$=100\\
 BFVGGSA  & $\alpha$=10, independent of a fixed $G_0$\\
 BPTGSA & $G_0$=100, $\alpha \in [5, 70]$ \\
 BNAGGSA & independent of $\alpha$ and $G_0$ both, $\delta$=$10^{-2}$, $\gamma$=$10^{-5}$ 
\\ 
\bottomrule
\end{longtable}
 \end{small}
 \end{center}

\begin{figure}[htbp]
\centering
\includegraphics[scale=0.4]{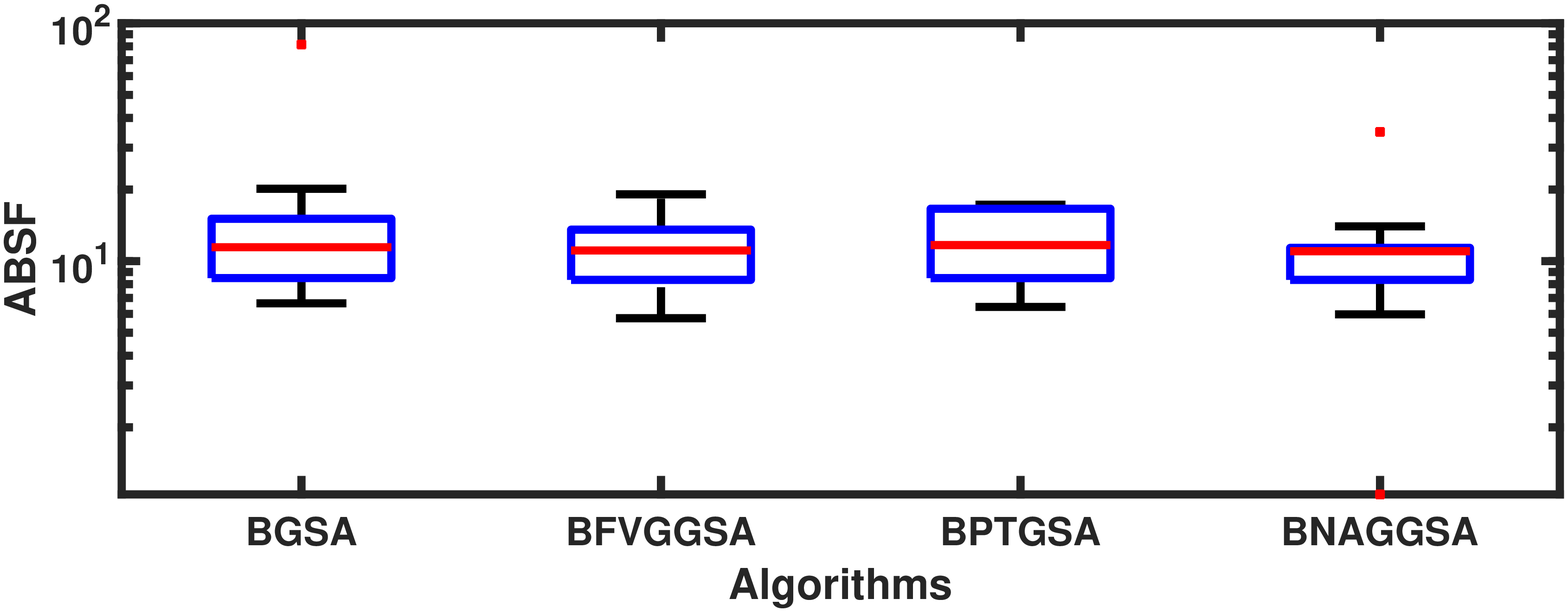}
    \caption{Boxplot analysis of average best so far ($ABSF$)} 
    \label{fig:boxplot_overall}    
\end{figure}

In Figure \ref{fig:boxplot_overall}, the boxplot of BNAGGSA have a less interquartile range and median as compared to BGSA, BFVGGSA and BPTGSA, which further implies that BNAGGSA is more efficient over other considered algorithms. To examine the significant difference among the results of the considered algorithms, Friedman test is used. This non-parametric statistical test is performed pairwise at 1\% level of significance with the null hypothesis, `There is no significant difference between the results obtained by the considered pair'. In this study, for pairwise comparisons, the adjusted p-values are reported. These p-values are achieved by the post-hoc test procedure, namely `bonferroni'. This procedure is implemented in R programming language \cite{ritu1,ritu2}. Table \ref{Ch5:Friedman_testbed5} presents the p-values for the pairwise comparison of BNAGGSA and other considered GSA binary variants. From Table \ref{Ch5:Friedman_testbed5}, the following observations are made:

\begin{itemize}
\item For the pair BFVGGSA-BNAGGSA, all the p-values are less than $0.01$ which implies that there is a significant difference between BFVGGSA-BNAGGSA over all the considered test problems. 
  
\item Except $f_{14}$ and $f_{18}$, all the p-values are less than $0.01$ for the  pair BGSA-BFVGGSA. It means, for $21$ problems ($f_1$, $f_2$, $f_3$, $f_4$, $f_5$, $f_6$, $f_7$, $f_8$, $f_9$, $f_{10}$, $f_{11}$, $f_{12}$, $f_{13}$, $f_{15}$, $f_{16}$, $f_{17}$, $f_{19}$, $f_{20}$, $f_{21}$, $f_{22}$ and $f_{23}$), there is a significant difference between BGSA-BFVGGSA. For $f_{14}$ and $f_{18}$, the p-values are greater than $0.01$, which indicates that BGSA performs similar to the BFVGGSA and vice versa.

\item  Except $f_3$, $f_5$, $f_{12}$, $f_{14}$, $f_{19}$ and $f_{22}$, all the p-values are less than $0.01$ for the pair BGSA-BPTGSA. It means, for $17$ problems ($f_1$, $f_2$, $f_4$, $f_6$, $f_7$, $f_8$, $f_9$, $f_{10}$, $f_{11}$,  $f_{13}$, $f_{15}$, $f_{16}$, $f_{17}$, $f_{18}$, $f_{20}$, $f_{21}$ and $f_{23}$), there is a significant difference between BGSA-BPTGSA. For $f_3$, $f_5$, $f_{12}$, $f_{14}$, $f_{19}$ and $f_{22}$, the p-values are greater than $0.01$, which indicates that BGSA performs similar to the BPTGSA and vice versa.

\item Except $f_{13}$ and $f_{23}$, all the p-values are less than $0.01$ for the  pair BGSA-BNAGGSA. It means, for $21$ problems ($f_1$, $f_2$, $f_3$, $f_4$, $f_5$, $f_6$, $f_7$, $f_8$, $f_9$, $f_{10}$, $f_{11}$, $f_{12}$, $f_{14}$, $f_{15}$, $f_{16}$, $f_{17}$, $f_{18}$, $f_{19}$, $f_{20}$, $f_{21}$ and $f_{22}$), there is a significant difference between BGSA-BNAGGSA. For $f_{13}$ and $f_{23}$, the p-values are greater than $0.01$, which indicates that BGSA performs similar to the BNAGGSA and vice versa.

\item  Except $f_6$, $f_7$ and $f_{14}$, all the p-values are less than $0.01$ for  the pair BFVGGSA-BPTGSA. It means, for $20$ problems ($f_1$, $f_2$, $f_3$, $f_4$, $f_5$, $f_8$, $f_9$, $f_{10}$, $f_{11}$, $f_{12}$, $f_{13}$, $f_{15}$, $f_{16}$, $f_{17}$, $f_{18}$, $f_{19}$, $f_{20}$, $f_{21}$, $f_{22}$ and $f_{23}$), there is a significant difference between BFVGGSA-BPTGSA. For $f_6$, $f_7$ and $f_{14}$, the p-values are greater than $0.01$, which indicates that BFVGGSA performs similar to the BPTGSA and vice versa. 

\item  Except $f_5$, $f_8$, $f_{15}$, $f_{16}$, $f_{20}$, $f_{22}$ and $f_{23}$, all the p-values are less than $0.01$ for the pair BPTGSA-BNAGGSA. It means, for $16$ problems ($f_1$, $f_2$, $f_4$, $f_6$, $f_7$, $f_9$, $f_{10}$, $f_{11}$, $f_{12}$, $f_{13}$, $f_{14}$, $f_{17}$, $f_{18}$, $f_{19}$ and $f_{21}$), there is a significant difference between BPTGSA-BNAGGSA. For $f_5$, $f_8$, $f_{15}$, $f_{16}$, $f_{20}$, $f_{22}$ and $f_{23}$, the p-values are greater than $0.01$, which indicates that BNAGGSA performs similar to the BPTGSA and vice versa.

\end{itemize}
Based on the above multiple comparative analyses, the proposed BNAGGSA is an overall better algorithm than other considered binary algorithms.

\begin{center}
\begin{small}
\setlength{\LTleft}{-20cm plus -1fill}
\setlength{\LTright}{\LTleft}
\begin{longtable}{p{1.5cm}p{2.0cm}p{2.5cm}p{2.5cm}p{2.5cm}p{2.5cm}}
 \caption{Performance of considered algorithms}
 \label{table:results}
\endfirsthead
\caption* {\textbf{Table \ref{table:results} Continued:}}
\endhead
\toprule
\textbf{Test problem} & \textbf{metrics} &  \textbf{BGSA}&  \textbf{BFVGGSA}  & 	\textbf{BPTGSA}  & 	\textbf{BNAGGSA} \\
 \midrule

 \multirow{3}{*} {$f_1$}			&	ABSF	&	3.086914619	&	0.067298611	&	3.103842338	&	\textbf{0.000352661}	\\	&	STDV	&	8.206405459	&	0.163922163	&	3.526355167	&	\textbf{0.00017413}	\\	&	Best	&	0.015050173	&	0	&	0.08687377	&	\textbf{0}	\\[1ex]
 
 \multirow{3}{*} {$f_2$}			&	ABSF	&	0.07792155	&	0.018473307	&	0.681757555	&	\textbf{0.003295898}	\\	&	STDV	&	0.047807411	&	0.016113692	&	0.421837265	&	\textbf{0.00161176}	\\	&	Best	&	0.007324219	&	0.001220703	&	0.042724609	&	\textbf{0}	\\[1ex]
 
 \multirow{3}{*} {$f_3$}			&	ABSF	&	70.47568758	&	94.49648857	&	148.2791752	&	\textbf{0.000362595}	\\	&	STDV	&	118.8177852	&	117.0452651	&	161.8985926	&	\textbf{7.38406E-05}	\\	&	Best	&	0.092685223	&	0.025033951	&	4.797875881	&	\textbf{0.000298023}	\\[1ex]
 
 \multirow{3}{*} {$f_4$}			&	ABSF	&	1.855875651	&	0.534667969	&	6.274007161	&	\textbf{0.030924479}	\\	&	STDV	&	1.428978817	&	0.690766982	&	7.254831838	&	\textbf{0.042137349}	\\	&	Best	&	0.390625	&	0.012207031	&	0.366210938	&	\textbf{0.012207031}	\\[1ex]
 
 \multirow{3}{*} {$f_5$}			&	ABSF	&	177.6295231	&	866.4641931	&	11352.02261	&	\textbf{23.95170429}	\\	&	STDV	&	572.0141956	&	1892.320518	&	33035.24801	&	\textbf{79.6930348}	\\	&	Best	&	3.516137163	&	0.581174682	&	3.890942624	&	\textbf{1.04698025}	\\[1ex]
 
 \multirow{3}{*} {$f_6$}			&	ABSF	&	4.4	&	0.1	&	4.5	&	\textbf{0}	\\	&	STDV	&	11.42978565	&	0.3	&	10.16120072	&	\textbf{0}	\\	&	Best	&	\textbf{0}	&	\textbf{0}	&	\textbf{0}	&	\textbf{0}	\\[1ex]
 

 \multirow{3}{*} {$f_7$}			&	ABSF	&	0.010974215	&	\textbf{0.002879371}	&	0.016518962	&	0.003187232	\\	&	STDV	&	0.006696426	&	0.002419516	&	0.014461618	&	\textbf{0.001758782}	\\	&	Best	&	0.001944507	&	0.00093562	&	0.002346033	&	\textbf{0.000877792}	\\[1ex]

 \multirow{3}{*} {$f_8$}			&	ABSF	&	-5390.732827	&	-5224.073913	&	-5310.475213	&	\textbf{-5589.778059}	\\	&	STDV	&	251.699878	&	250.1758352	&	238.8942127	&	\textbf{136.6408955}	\\	&	Best	&	-5806.389527	&	-5636.886076	&	-5804.293339	&	\textbf{-6037.440855}	\\[1ex]
 
 \multirow{3}{*} {$f_9$}			&	ABSF	&	8.134128779	&	4.404034452	&	5.975501163	&	\textbf{2.728963629}	\\	&	STDV	&	3.665399353	&	1.803598251	&	2.295928528	&	\textbf{1.502279429}	\\	&	Best	&	3.241000492	&	0.003332237	&	2.450572396	&	\textbf{0.00023249}	\\[1ex]

 \multirow{3}{*} {$f_{10}$}			&	ABSF	&	4.865229489	&	0.262288777	&	0.438199742	&	\textbf{0.00950132}	\\	&	STDV	&	1.754725389	&	0.619683163	&	0.737461356	&	\textbf{0.004531467}	\\	&	Best	&	2.396765215	&	0.007150226	&	0.007150226	&	\textbf{8.88178E-16}	\\[1ex]

 \multirow{3}{*} {$f_{11}$}			&	ABSF	&	0.437071984	&	0.086784975	&	0.95415136	&	\textbf{0.047908036}	\\	&	STDV	&	0.257320344	&	0.057438425	&	0.446455815	&	\textbf{0.023087272}	\\	&	Best	&	0.068297739	&	0.017723767	&	0.257408952	&	\textbf{0.011917058}	\\[1ex]

 \bottomrule
 \pagebreak
\toprule
\textbf{Test problem} & \textbf{metrics} &  \textbf{BGSA}&  \textbf{BFVGGSA}  & 	\textbf{BPTGSA}  & 	\textbf{BNAGGSA} \\
 \midrule

 \multirow{3}{*} {$f_{12}$}			&	ABSF	&	1.801869755	&	5.751197856	&	1592.485357	&	\textbf{0.895785387}	\\	&	STDV	&	1.875365125	&	12.73916346	&	7684.463846	&	\textbf{0.985715835}	\\	&	Best	&	0.002877555	&	0.000383423	&	0.141754486	&	\textbf{3.88349E-06}	\\[1ex]

 \multirow{3}{*} {$f_{13}$}			&	ABSF	&	9.151127448	&	8.096012009	&	24728.1539	&	\textbf{0.285255839}	\\	&	STDV	&	45.84203549	&	37.69028395	&	126261.0999	&	\textbf{0.120808877}	\\	&	Best	&	0.030214301	&	\textbf{0.004561255}	&	0.141504394	&	0.101241856	\\[1ex]
 
 \multirow{3}{*} {$f_{14}$}			&	ABSF	&	\textbf{1.02513438}	&	1.0327565	&	1.239623931	&	1.06814383	\\	&	STDV	&	\textbf{0.064307845}	&	0.07774098	&	0.37739433	&	0.118627136	\\	&	Best	&	0.998003842	&	\textbf{0.998003838}	&	\textbf{0.998003838}	&	\textbf{0.998003838}	\\[1ex]

 \multirow{3}{*} {$f_{15}$}			&	ABSF	&	0.001541094	&	0.002879668	&	0.003562143	&	\textbf{0.001448739}	\\	&	STDV	&	0.000233295	&	0.000978373	&	0.001572025	&	\textbf{0.000212638}	\\	&	Best	&	0.001142737	&	0.001501691	&	0.000943598	&	\textbf{0.001007221}	\\[1ex]

 \multirow{3}{*} {$f_{16}$}			&	ABSF	&	-1.030965569	&	-1.024444516	&	-1.025333604	&	\textbf{-1.031627651}	\\	&	STDV	&	0.001769449	&	0.009623901	&	0.008259233	&	\textbf{8.1049E-07}	\\	&	Best	&	\textbf{-1.031627919}	&	-1.031627919	&	-1.031627284	&	\textbf{-1.031627919}	\\[1ex]
 

 \multirow{3}{*} {$f_{17}$}			&	ABSF	&	0.398881909	&	0.39955722	&	0.401005955	&	\textbf{0.397892311}	\\	&	STDV	&	0.003392942	&	0.003648867	&	0.005111368	&	\textbf{1.30556E-05}	\\	&	Best	&	\textbf{0.397887483}	&	0.397887838	&	0.397888021	&	0.39788759	\\[1ex]

 \multirow{3}{*} {$f_{18}$}			&	ABSF	&	3.000014948	&	3.071615699	&	3.102829593	&	\textbf{3.000014448}	\\	&	STDV	&	3.612E-06	&	0.111115908	&	0.151152665	&	\textbf{2.31263E-06}	\\	&	Best	&	\textbf{3}	&	\textbf{3}	&	\textbf{3}	&	\textbf{3}	\\[1ex]
 
 \multirow{3}{*} {$f_{19}$}			&	ABSF	&	-3.862565394	&	-3.862497385	&	-3.862497079	&	\textbf{-3.862673046}	\\	&	STDV	&	0.000484628	&	0.000546308	&	0.000554687	&	\textbf{0.000407993}	\\	&	Best	&	-3.862781675	&	\textbf{-3.862782133}	&	-3.862781161	&	\textbf{-3.862782133}	\\[1ex]

 \multirow{3}{*} {$f_{20}$}			&	ABSF	&	-3.009940986	&	-3.20643504	&	\textbf{-3.213241383}	&	-3.202788907	\\	&	STDV	&	0.163169912	&	0.047910634	&	0.034694584	&	\textbf{0.006513749}	\\	&	Best	&	-3.253592834	&	\textbf{-3.32192263}	&	-3.318878429	&	-3.235974116	\\[1ex]

 \multirow{3}{*} {$f_{21}$}			&	ABSF	&	-3.246885805	&	-3.454377467	&	-3.017673843	&	\textbf{-3.65082142}	\\	&	STDV	&	1.923145867	&	1.8907371	&	\textbf{0.725104293}	&	1.42087111	\\	&	Best	&	-10.08344751	&	\textbf{-10.15318865}	&	-5.098486745	&	\textbf{-10.15318865}	\\[1ex]
 
 \multirow{3}{*} {$f_{22}$}			&	ABSF	&	-4.356449686	&	-5.246731263	&	-4.58125434	&	\textbf{-9.955602484}	\\	&	STDV	&	2.592617445	&	3.387246827	&	2.938935908	&	\textbf{1.665479577}	\\	&	Best	&	-10.376123	&	\textbf{-10.40292776}	&	-10.40290812	&	\textbf{-10.40292776}	\\[1ex]
 
 \multirow{3}{*} {$f_{23}$}			&	ABSF	&	-4.325671493	&	-3.684812264	&	-3.38391909	&	\textbf{-5.025986572}	\\	&	STDV	&	2.816390967	&	2.713889997	&	\textbf{2.094425651}	&	2.991699804	\\	&	Best	&	-10.21916799	&	-10.53621447	&	\textbf{-10.53640018}	&	\textbf{-10.53640018}	\\[1ex]

\bottomrule
\end{longtable}
 \end{small}
 \end{center}

\begin{center}
\begin{small}
\setlength{\LTleft}{-20cm plus -1fill}
\setlength{\LTright}{\LTleft}
\begin{longtable}{p{1cm}p{2cm}p{2.5cm}p{2.5cm}p{2.5cm}}
 \caption{p-values for pairwise comparison of considered algorithms}
 \label{Ch5:Friedman_testbed5}
\endfirsthead
\caption* {\textbf{Table \ref{Ch5:Friedman_testbed5} Continued:}}
\endhead
\toprule
\textbf{TP } & \textbf{-} &  \textbf{BGSA}&  \textbf{BFVGGSA}  &	\textbf{BPTGSA}  \\
 \midrule

 \multirow{3}{*} {$f_{1}$}	&	BFVGGSA	&	$<2E-16$ & $-$  & $-$ 		\\	& BPTGSA	&	$<2E-16$ & $<2E-16$  & $-$ 	  	\\ & BNAGGSA	&	$7.9E-12$ & $<2E-16$ & $<2E-16$   \\ [1ex]

 \multirow{3}{*} {$f_{2}$}	&	BFVGGSA	&	$<2E-16$ & $-$  & $-$ 		\\	& BPTGSA	&	$<2E-16$ & $<2E-16$  & $-$ 	  	\\ & BNAGGSA	&	$<2E-16$ & $<2E-16$ & $<2E-16$   \\ [1ex]

       \bottomrule
\pagebreak         
\toprule
\textbf{TP } & \textbf{-} &  \textbf{BGSA}&  \textbf{BFVGGSA}  &	\textbf{BPTGSA}  \\
 \midrule

  \multirow{3}{*} {$f_{3}$}&	BFVGGSA	&	$<2E-16$ & $-$  & $-$ 		\\	& BPTGSA	&	$<1$ & $<2E-16$  & $-$ 	  	\\ & BNAGGSA	&	$1.6E-11$ & $<2E-16$ & $1.6E-11$   \\ [1ex]

   \multirow{3}{*} {$f_{4}$}	&	BFVGGSA	&	$<2E-16$ & $-$  & $-$ 		\\	& BPTGSA	&	$<2E-16$ & $<2E-16$  & $-$ 	  	\\ & BNAGGSA	&	$ 5.7E-08$ & $<2E-16$ & $<2E-16$   \\ [1ex]

    \multirow{3}{*} {$f_{5}$}	&	BFVGGSA	&	$<2E-16$ & $-$  & $-$ 		\\	& BPTGSA	&	$0.012$ & $<2E-16$  & $-$ 	  	\\ & BNAGGSA	&	$0.167$ & $<2E-16$ & $1.00$   \\ [1ex]

     \multirow{3}{*} {$f_{6}$}	&	BFVGGSA	&	$<2E-16$ & $-$  & $-$ 		\\	& BPTGSA	&	$<2E-16$ & $0.34$  & $-$ 	  	\\ & BNAGGSA	&	$1E-07$ & $<2E-16$ & $<2E-16$   \\ [1ex]

      \multirow{3}{*} {$f_{7}$}&	BFVGGSA	&	$<2E-16$ & $-$  & $-$ 		\\	& BPTGSA	&	$<2E-16$ & $1$  & $-$ 	  	\\ & BNAGGSA	&	$0.0088$ & $<2E-16$ & $<2E-16$   \\ [1ex]

       \multirow{3}{*} {$f_{8}$}&	BFVGGSA	&	$ 2.1E-10$ & $-$  & $-$ 		\\	& BPTGSA	&	$4.8E-11$ & $<2E-16$  & $-$ 	  	\\ & BNAGGSA	&	$ 9.0E-10$ & $<2E-16$ & $1$   \\ [1ex]

        \multirow{3}{*} {$f_{9}$}&	BFVGGSA	&	$<2E-16$ & $-$  & $-$ 		\\	& BPTGSA	&	$<2E-16$ & $1.3E-11$  & $-$ 	  	\\ & BNAGGSA	&	$ 1.1E-07$ & $<2E-16$ & $ 2.6E-05$   \\ [1ex]

         \multirow{3}{*} {$f_{10}$}	&	BFVGGSA	&	$<2E-16$ & $-$  & $-$ 		\\	& BPTGSA	&	$<2E-16$ & $<2E-16$  & $-$ 	  	\\ & BNAGGSA	&	$<2E-16$ & $<2E-16$ & $1.2E-07$   \\ [1ex]

          \multirow{3}{*} {$f_{11}$}	&	BFVGGSA	&	$<2E-16$ & $-$  & $-$ 		\\	& BPTGSA	&	$<2E-16$ & $0.00053$  & $-$ 	  	\\ & BNAGGSA	&	$<2E-16$ & $<2E-16$ & $<2E-16$   \\ [1ex]

           \multirow{3}{*} {$f_{12}$}&	BFVGGSA	&	$3.4E-06$ & $-$  & $-$ 		\\	& BPTGSA	&	$0.82$ & $4.8E-09 $  & $-$ 	  	\\ & BNAGGSA	&	$1.1E-12$ & $<2E-16$ & $1.2E-09$   \\ [1ex]

            \multirow{3}{*} {$f_{13}$}	&	BFVGGSA	&	$<2E-16$ & $-$  & $-$ 		\\	& BPTGSA	&	$0.00056$ & $<1E-08$  & $-$ 	  	\\ & BNAGGSA	&	$1$ & $<2E-16$ & $0.00158$   \\ [1ex]

             \multirow{3}{*} {$f_{14}$}	&	BFVGGSA	&	$1$ & $-$  & $-$ 		\\	& BPTGSA	&	$1$ & $1$  & $-$ 	  	\\ & BNAGGSA	&	$0.00226$ & $6.8E-05$ & $0.00032$   \\ [1ex]

              \multirow{3}{*} {$f_{15}$}&	BFVGGSA	&	$0.045$ & $-$  & $-$ 		\\	& BPTGSA	&	$<2E-16$ & $<2E-16$  & $-$ 	  	\\ & BNAGGSA	&	$<2E-16$ & $<2E-16$ & $1$   \\ [1ex]

               \multirow{3}{*} {$f_{16}$}	&	BFVGGSA	&	$6.7E-13$ & $-$  & $-$ 		\\	& BPTGSA	&	$<2E-16$ & $<2E-16$  & $-$ 	  	\\ & BNAGGSA	&	$<2E-16$ & $<2E-16$ & $1$   \\ [1ex]

                \multirow{3}{*} {$f_{17}$}&	BFVGGSA	&	$0.00038$ & $-$  & $-$ 		\\	& BPTGSA	&	$8.7E-13$ & $<2E-16$  & $-$ 	  	\\ & BNAGGSA	&	$<2E-16$ & $<2E-16$ & $9.8E-08$   \\ [1ex]

                 \multirow{3}{*} {$f_{18}$}&	BFVGGSA	&	$1$ & $-$  & $-$ 		\\	& BPTGSA	&	$<2E-16$ & $8E-16$  & $-$ 	  	\\ & BNAGGSA	&	$<2E-16$ & $<2E-16$ & $0.00057$   \\ [1ex]

            \bottomrule
\pagebreak         
\toprule
\textbf{TP } & \textbf{-} &  \textbf{BGSA}&  \textbf{BFVGGSA}  &	\textbf{BPTGSA}  \\
 \midrule

                  \multirow{3}{*} {$f_{19}$}	&	BFVGGSA	&	$<2E-16$ & $-$  & $-$ 		\\	& BPTGSA	&	$<1$ & $<2E-16$  & $-$ 	  	\\ & BNAGGSA	&	$0.00046$ & $<2E-16$ & $0.00282$   \\ [1ex]

                   \multirow{3}{*} {$f_{20}$}&	BFVGGSA	&	$<2E-16$ & $-$  & $-$ 		\\	& BPTGSA	&	$<2E-16$ & $4.6E-06$  & $-$ 	  	\\ & BNAGGSA	&	$<2E-16$ & $0.216 $ & $0.012$   \\ [1ex]

                    \multirow{3}{*} {$f_{21}$}	&	BFVGGSA	&	$<2E-16$ & $-$  & $-$ 		\\	& BPTGSA	&	$1.7E-10$ & $6.8E-10 $  & $-$ 	  	\\ & BNAGGSA	&	$0.00029$ & $<2E-16$ & $0.00701$   \\ [1ex]

                     \multirow{3}{*} {$f_{22}$}&	BFVGGSA	&	$<2E-16$ & $-$  & $-$ 		\\	& BPTGSA	&	$9.7E-08$ & $<2E-16$  & $-$ 	  	\\ & BNAGGSA	&	$7.7E-05$ & $<2E-16$ & $0.68  $   \\ [1ex]

                      \multirow{3}{*} {$f_{23}$}	&	BFVGGSA	&	$1.5E-09$ & $-$  & $-$ 		\\	& BPTGSA	&	$0.29$ & $1.2E-13$  & $-$ 	  	\\ & BNAGGSA	&	$0.54$ & $ 4.8E-13$ & $1$   \\ [1ex]

\bottomrule
\end{longtable}
 \end{small}
 \end{center}

\section{Windfarm Layout Optimization Problem}\label{sec:windfarm}

Wind energy is one of the fast growing and most sustainable resource of the renewable energy in which the wind power is extracted by the turbine through rotation of its blades. The installation of the turbines in the windfarm is the most important entity as the wake loss induced by upstream turbines affect the output of downstream turbines, which further reduces the total power output of the windfarm \cite{biswas2018decomposition}. Therefore, to maximize the total output of the windfarm the optimal pattern of wind turbines is required subject to the constraints related to the position of the turbines, rotor radius and farm radius. In \cite{biswas2018decomposition}, a wind model with real data sets is optimized under the decomposition based multi-objective environment. In this study, the practical applicability of the proposed variant BNAGGSA is proven to optimize the wind model of \cite{biswas2018decomposition} under the single objective environment, in the binary search space.

\subsection{Mathematical models of the considered windfarm and its numerical data}
\begin{figure}[H]
 \subfigure[$k^{th}$ turbine under the wake effect produced by $m^{th}$ turbine]{
   \includegraphics[height=4.5cm, width=6.6cm]{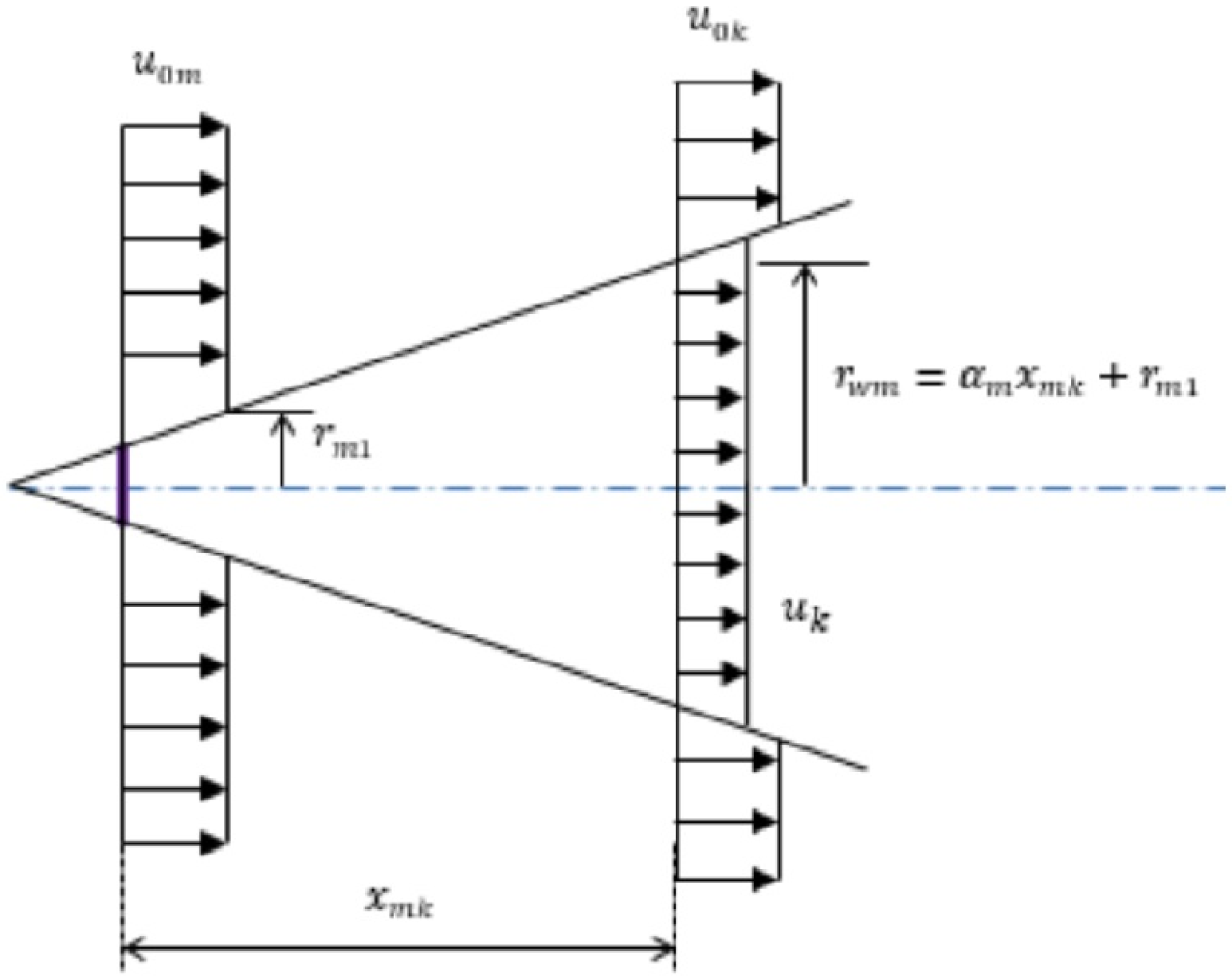}}
\label{fig:wake_effect1} 
 \subfigure[$i^{th}$ turbine partially influenced by wake radius $r_{wj}$ of $j^{th}$ turbine]{
    \includegraphics[height=4.5cm, width=6.6cm]{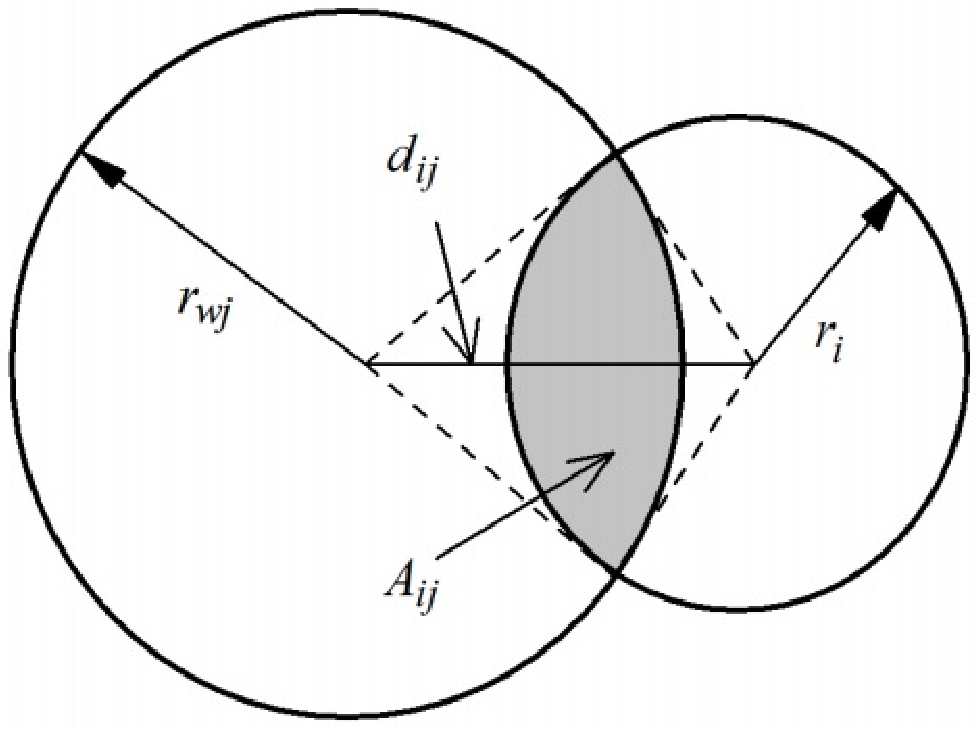}}
    \label{fig:wake_effect2}
  \caption{Graphical representation of Jensen wake decay model \cite{biswas2018decomposition} } 
  \label{fig:wake_effect}  
  \end{figure}

The windfarm layout design depends upon its minimum wake loss strategy. Wake is the only significant constraint which effects the wind model's objectives more drastically. Wake exists when wind stream interacts with turbines. This interaction not only reduces the speed of the wind but also increases the turbulence intensity of the attached turbines. Two more issues with the existed wake are its movement towards downstream direction and its diverging nature. Under the wake effects, turbines which are located in downstream regions do not perform well. In the past years, several wake models have been developed to reduce these wake effects on the performance of their respective windfarms. In this study, Jensen wake decay model \cite{biswas2018decomposition} is taken to find the wind velocity inside the wake region. Let us consider $k^{th}$ turbine under the wake region of the $m^{th}$ turbine (refer Figure \ref{fig:wake_effect}(a)). According to the  Jensen wake decay model, the wind speed for the $k^{th}$ turbine is calculated as:

\begin{equation}\label{eqn:WE1}
u_k=u_{0k}\Big[1-\frac{2a}{(1+\alpha_m\frac{x_{mk}}{r_{m1}})}\Big]
\end{equation}

\begin{equation}\label{eqn:WE2}
a=\frac{1-\sqrt{1-C_T}}{2}
\end{equation}

\begin{equation}\label{eqn:WE3}
r_{m1}=r_m\sqrt{\frac{1-a}{1-2a}}
\end{equation}

\begin{equation}\label{eqn:WE4}
\alpha_m=\frac{0.5}{ln(\frac{h_m}{z_0})}
\end{equation}

where,
\begin{itemize}
 \item $u_{ok}$ is the local wind speed at $k^{th}$ turbine without considering the wake effect, 
  \item  $x_{mk}$ is the distance between $m^{th}$ and $k^{th}$ turbine,
 \item  $ r_m$ is the radius of $m^{th}$ turbine rotor, 
 \item  $r_{m1}$ is the downstream rotor radius of $m^{th}$ turbine, 
 \item  $h_m$ is the hub height of the $m^{th}$turbine,
 \item  $\alpha_m$ is the entrainment constant pertaining to $m^{th}$ turbine, 
 \item  $a$ is the axial induction factor, 
 \item   $C_T$ is the thrust coefficient of the wind turbine rotor, 
  \item  $z_0$ is the surface roughness of the windfarm.
\end{itemize}

The wake region of a linear wake model is conical with the wake influence radius defined as:
\begin{equation}\label{eqn:WE5}
r_{wm}=\alpha_mx_{mk}+r_{m1}
\end{equation}

Since the velocity of wind is different for different heights therefore the local wind speed at $k^{th}$ turbine depends upon its hub hight defined as:

\begin{equation}\label{eqn:WE6}
u_{0k}=u_{ref}log(\frac{h_k}{z_0})/log(\frac{h_{ref}}{z_0})
\end{equation}

where, $u_{ref}$ is the wind speed at the reference height $h_{ref}$. If the $i^{th}$ turbine is inside the area of multiple wake flows, than the wind speed at that turbine can be calculated as:

\begin{equation}\label{eqn:WE7}
u_i=u_{0i}\Big[1-\sqrt{\sum_{j=1}^{N_t}\frac{A_{ij}}{r_i^2}(1-\frac{u_{ij}}{u_{0j}})^2}\Big]
\end{equation}

where, $u_{0i}$ and $u_{0j}$ are the local wind speeds at $i^{th}$ and $j^{th}$ turbines, respectively without considering the wake effect; $u_{ij}$  is the wind velocity at $i^{th}$ turbine under the influence of $j^{th}$ turbine; $N_t$ is the number of turbines affecting the $i^{th}$ turbine with wake effects; $r_i$ is the rotor radius of $i^{th}$ turbine; $A_{ij}$ is the overlapped rotor area of $i^{th}$ turbine under wake influence radius ($r_{wj}$) of $j^{th}$ turbine (Figure \ref{fig:wake_effect}(b)). Under the condition of full wake effect, $A_{ij}=\pi \times r_i^2$ \cite{biswas2018decomposition}.

\subsubsection{Turbine power output model}

The output power (in kW) from the turbine is a function of wind speed (u) and expressed as:
\begin{equation}\label{eqn:WE8}
P(u) = \begin{cases}
0, &\text{for $u<u_c$}\\
60*(u-u_c) &\text{for $u_c \leq u \leq u_c+\frac{25}{19}$}\\
250*(u-u_c-1) &\text{for $u_c+\frac{25}{19} \leq u \leq 13$}\\
2000 &\text{for $13 \leq u \leq u_f$}\\
0 &\text{for $ u>u_f$}\\
\end{cases}
\end{equation}
where, $u_c$ and $u_f$ are the cut-in and cut-out wind speeds of the turbine, respectively.

\subsubsection{Wind condition model}
\begin{figure}
 \subfigure[Wind probability distribution diagram for site 1]{
    \includegraphics[height=3.5cm, width=12cm]{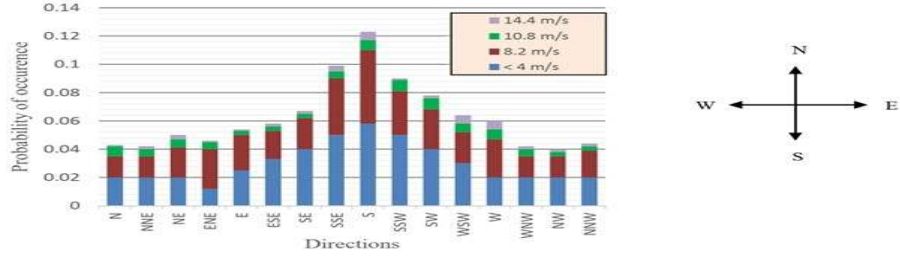}}
     \label{fig:wind1}
\subfigure[Wind probability distribution diagram for site 2]{
     \includegraphics[height=3.5cm, width=12cm]{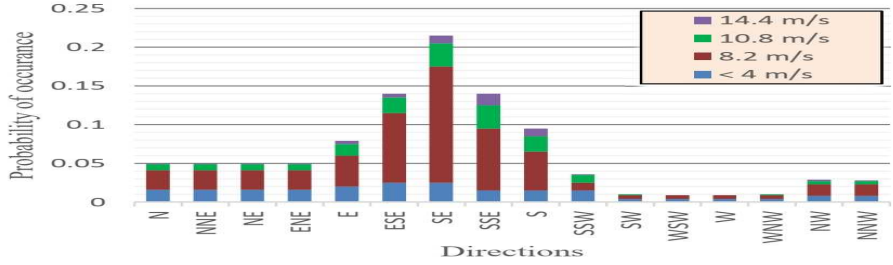}}
    \label{fig:wind2}
  \caption{Wind probability distribution diagram for both sites \cite{biswas2018decomposition}}   
 \label{fig:winddata}
  \end{figure}

 In \cite{biswas2018decomposition}, the authors considered two type of wind data sets of sites in Midland (site $1$) and Corpus Christi (site $2$), Texas. The wind probability distribution diagrams for site $1$ and $2$ are presented in Figure \ref{fig:winddata}(a) and Figure \ref{fig:winddata}(b), respectively. The direction of the each wind data is discretized into $16$ different segments of $22.5^{\circ}$ each. The north direction set as $0^{\circ}$ and an increment of $22.5^{\circ}$ in clockwise direction is considered to divide $360^{\circ}$ in $16$ segments \cite{biswas2018decomposition}. Wind speed is discretized into these $16$ segments in such a way that each segment have the probability of occurrence of each discrete wind speed.  The summation of all the probabilities equals $1$. For the both case studies, four discrete wind speeds are considered as $u_{ref_0}< 4m/s$, $u_{ref_1}<8.2m/s$, $u_{ref_2} < 10.8m/s$ and $u_{ref_4}= 14.4m/s$ at hub height= $60m$ \cite{biswas2018decomposition}. In this study, both the mentioned cases are reconsidered to optimize their respective objectives more accurately. 

\subsubsection{Objective function under consideration}

In this study, two different objectives are maximized in the single objective platform through the weighted aggregate method \cite{marler2010weighted} written as:
\begin{equation}\label{eqn:WE9}
F=w_1 \times F_1+w_2 \times F_2
\end{equation}
Where, $F_1$ and $F_2$ are the efficiency and the total power output of the windfarm calculated as follows:

\begin{equation}\label{eqn:WE10}
F_1=\frac{\sum_{k=0}^{360}\sum_{i=1}^{N_T}f_kP_i(u_i)}{\sum_{k=0}^{360}\sum_{i=1}^{N_t}f_kP_{i,max}(\bar{u_i})}
\end{equation}

Total power output of the windfarm is given by,
\begin{equation}\label{eqn:WE11}
F_2=\sum_{k=0}^{360}\sum_{i=1}^{N_T}f_kP_i(u_{i})
\end{equation}

where, $N_T$ is the total number of turbines, $P_{i,max}$ is the maximum power output from $i^{th}$ turbine with the wind speed $\bar{u_{l}}$ under no wake effect, $P_i$ is the actual power output from $i^{th}$ turbine as a function of wind speed $u_i$ considering wake effect it experiences from upstream turbine(s). The probability of occurrence of each wind speed from each direction is defined by factor $f_k$ and $\sum_{k=0}^{360}f_k=1$. $w_1$ and $w_2$ are weights to the objective functions $F_1$ and $F_2$ respectively and set to $0.5$.

\subsubsection{Windfarm selection and turbine data}

In this study, a rectangular windfarm of $6000m \times 2000m$ is divided into $100$ equal cells of size $300m \times 400m$. At the center of each cell, a turbine can be placed. The objective of this study is to maximize the total power output of this windfarm having $N_t$ number of turbines in which each turbine have $100$ possible locations to be placed. The turbine and other relevant data used for calculation are listed in Table \ref{table:turbine}.

 \begin{center}
\begin{small}
\setlength{\LTleft}{-20cm plus -1fill}
\setlength{\LTright}{\LTleft}
\begin{longtable}{p{5.7cm}p{2.5cm}}
 \caption{Turbine and other relevant data}
 \label{table:turbine}
\endfirsthead
\endhead
\toprule
\textbf{Parameter } & \textbf{Value}\\
 \midrule
 Turbine model & Vestas V-80 \\	
Rated Power, $P_{rated}$ & 2000 kW\\
Rotor diameter, $D$ & 80m\\															Thrust coefficient, $C_T$ & 0.8\\													Hub hight, (IEC IA) & 60m, 67m or 78m\\	
Cut-in speed, $u_c$ & 4m/s \\	
Cut-out speed, $u_f$ & 25m/s \\	
Surface roughness of windfarm, $z_0$ & 0.3m \\	
Reference height, $h_{ref}$ & 60m \\						
\bottomrule
\end{longtable}
 \end{small}
 \end{center}

\section{Applicability of the proposed variant BNAGGSA for wind farm layout
optimization problem} \label{Applicab}
GSA is a well established optimizer to solve the real world optimization problems in the various fields of engineering science, like power engineering, pattern recognition, image processing, classification, communication engineering, control engineering, civil engineering, computer and software engineering, mechanical engineering, water industry etc. As per the authors' best knowledge, no application work of GSA framework has been done so far in the field of renewable energy. In this study, GSA framework is used in the binary search space for optimizing the wind farm layout to obtain the maximum power output. 

If $N$ is the swarm size than there are $N$ candidate solutions $X_1$, $X_2$,...,$X_N$ for the considered windfarm layout optimization problem. Since each turbine has $100$ possible locations to be placed therefore each candidate solution ($X_k$ for $k=1,2,...,N$) is 100-dimensional vector where each dimension represents the status of that location through the binary values $0$ or $1$. $0$ indicates that the location is empty while $1$ indicates the presence of the turbine. Thus the $k^{th}$ candidate solution is encoded as binary vector  $X_k=(x_1, x_2,...,x_{100})$ where $x_j, \forall j=1:100$ is either $0$ or $1$.

In this study, we optimize different windfarm layout for different number of turbines ($N_T$) by considering it as a constraint optimization problem whose objection function $F$ (refer equation (\ref{eqn:WE9})) is re-defined using the penalty function approach as follows:
\begin{equation}
F(X_i) = \begin{cases}
F(X_i),&\text{if sum of ones in $X_i =N_T$}\\
P, &\text{otherwise}\\
\end{cases}
\end{equation}
Here $P$ is the penalty due to the constraints violation. Since this is a maximization problem therefore $P$ is assigned very small value set to $10^{-10}$. The different values of $N_T$ are $10$, $20$, $30$, $40$, $50$ and $60$ taken under consideration.

\subsection{Results and discussion}\label{result}
In this section, the proposed binary variant is tested over two different wind data sets of two different wind sites (site $1$ and $2$) for the considered windfarm. All the required information about both sites are given in section \ref{sec:windfarm}.  In both cases, the heights of all turbines are fixed. The wind deficit for a turbine under the wake effect is calculated by equations (\ref{eqn:WE1}) to (\ref{eqn:WE7}). The power output and the objective functions are calculated using equations (\ref{eqn:WE8}) to (\ref{eqn:WE11}). 

In order to validate the effectiveness and robustness of proposed algorithm, BNAGGSA is compared with binary GSA \cite{rashedi2010bgsa}, a binary version of 
FVGGSA \cite{Bansal2018}, binary version of PTGSA \cite{joshi2019parameter}
along with MOEA/D \cite{biswas2018decomposition}. The binary version of FVGGSA (BFGGSA) and PTGSA (BPTGSA) are followed the same settings as binary GSA (refer section \ref{sec:binary_version}). For all the comparison, the following experimental setting is adopted:

\begin{itemize}
\item Swarm size =$500$,
\item  Maximum number of generations/iterations =$500$,
\item Total number of runs/simulations =$10$,
\item The other parameters of BGSA, BNAGGSA, BFVGGSA, BPTGSA and MOEA/D are considered from their original resources  while the result of MOEA/D is reproduced from \cite{biswas2018decomposition}. 
\item The parameters of the consider windfarm and its turbines are already mentioned in section \ref{sec:windfarm}. 
\end{itemize}

The results of optimized layouts of site $1$ and site $2$ with different numbers of turbines ($N_T$) along with corresponding power output, efficiency and capacity factor are listed in Table \ref{table:state1} and Table \ref{table:state2}, respectively. Capacity factor is a ratio between the amount of energy delivered annually and the maximum amount of energy can be delivered yearly. It is clear from the results that the proposed BNAGGSA outperforms in every metric of comparison with BGSA, BFVGGSA and BPTGSA for both sites. For the optimal layout of wind site $1$, MOED/D is performed superior for $10$ and $20$ number of turbines. For the optimal layout of $30$, $40$, $50$ and $60$ number of turbines, the proposed variant proves its superimacy with MOED/D in each metric of comparison. For the wind site $2$, MOED/D performs better (in each metric) than the proposed BNAGGSA for the layout of $10$ and $20$ turbines. For $30$ turbines layout, MOED/D performs better than BNAGGSA only in power out and capacity factor while BNAGGSA is superior in terms of efficiency. For $40$, $50$ and $60$ turbines layout, the proposed BNAGGSA is superior than MOED/D in all the metrics of comparison. 

\begin{figure}
\centering
\subfigure[Optimal layout for $10$ turbines]{
   \includegraphics[height=3.1cm, width=11.6cm]{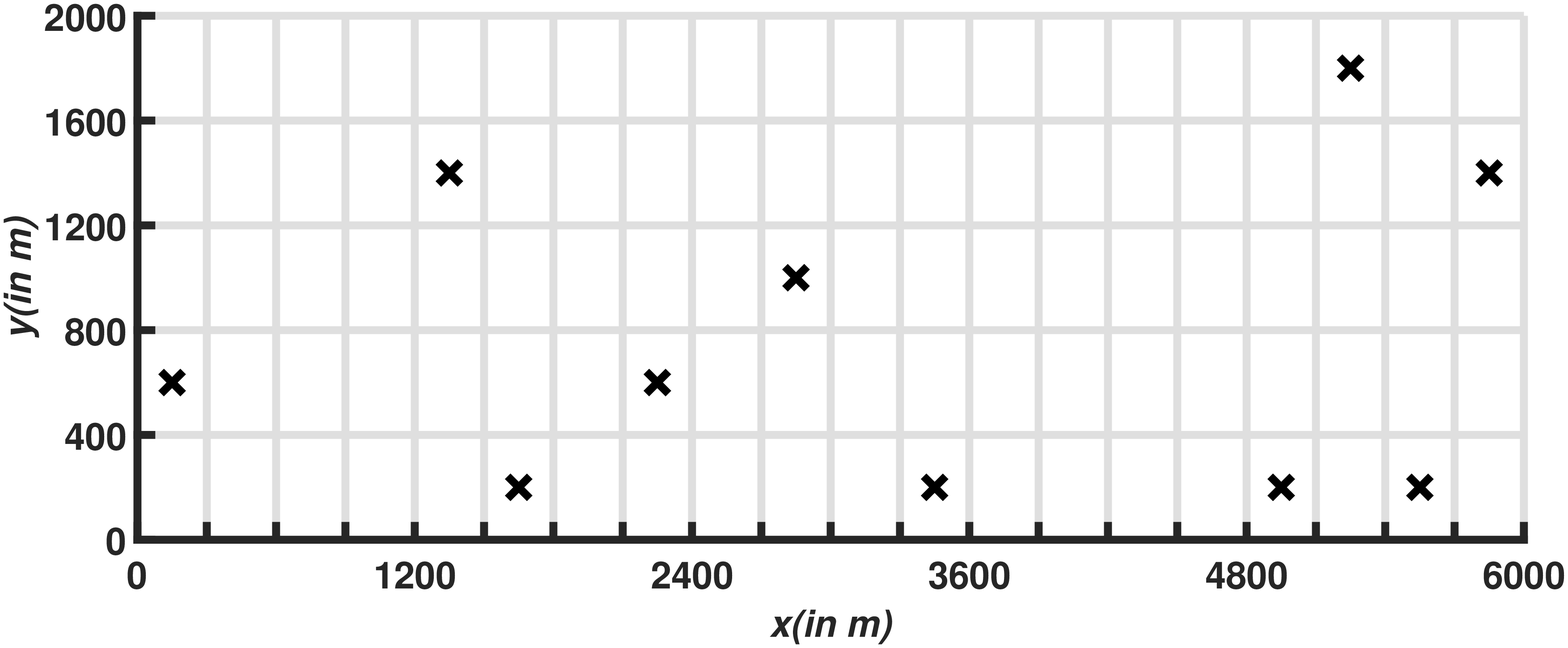}}
\subfigure[Optimal layout for $20$ turbines]{
  \includegraphics[height=3.1cm, width=11.6cm]{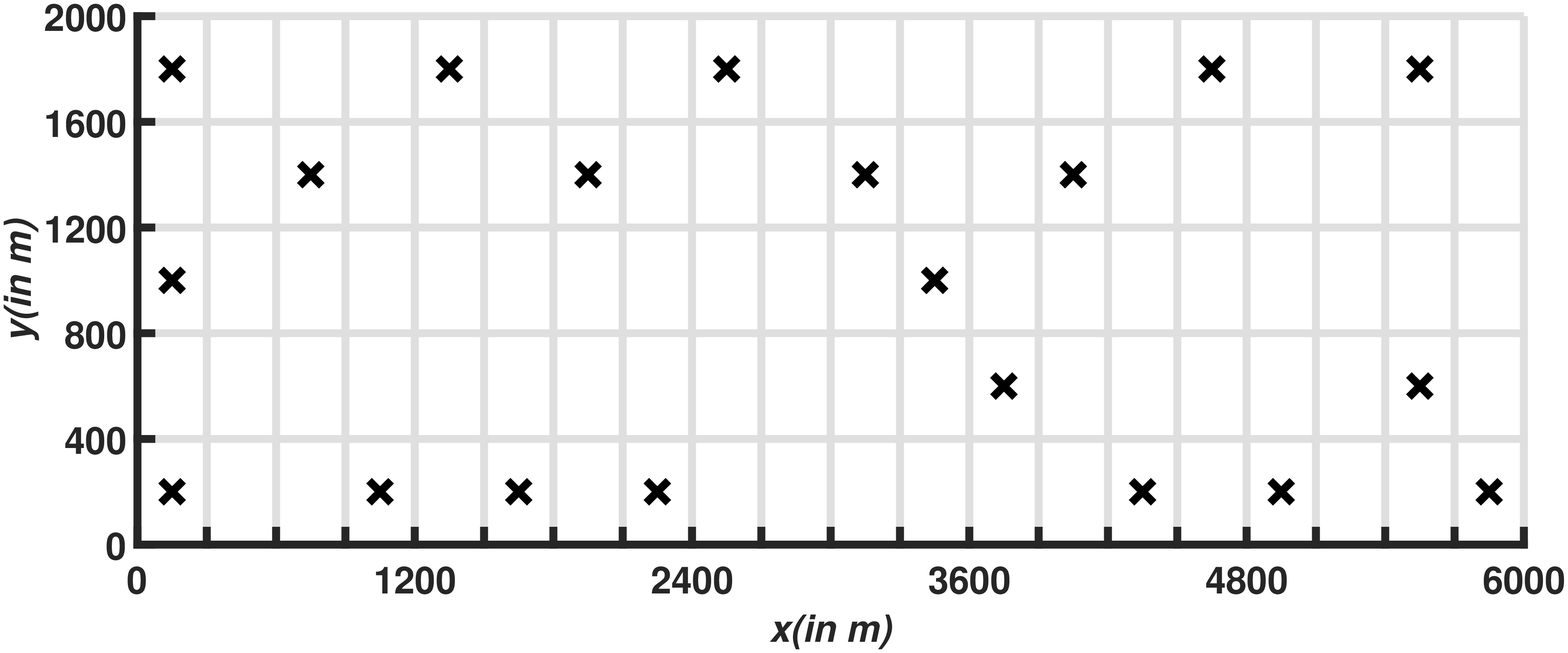}}
\subfigure[Optimal layout for $30$ turbines]{
   \includegraphics[height=3.1cm, width=11.6cm]{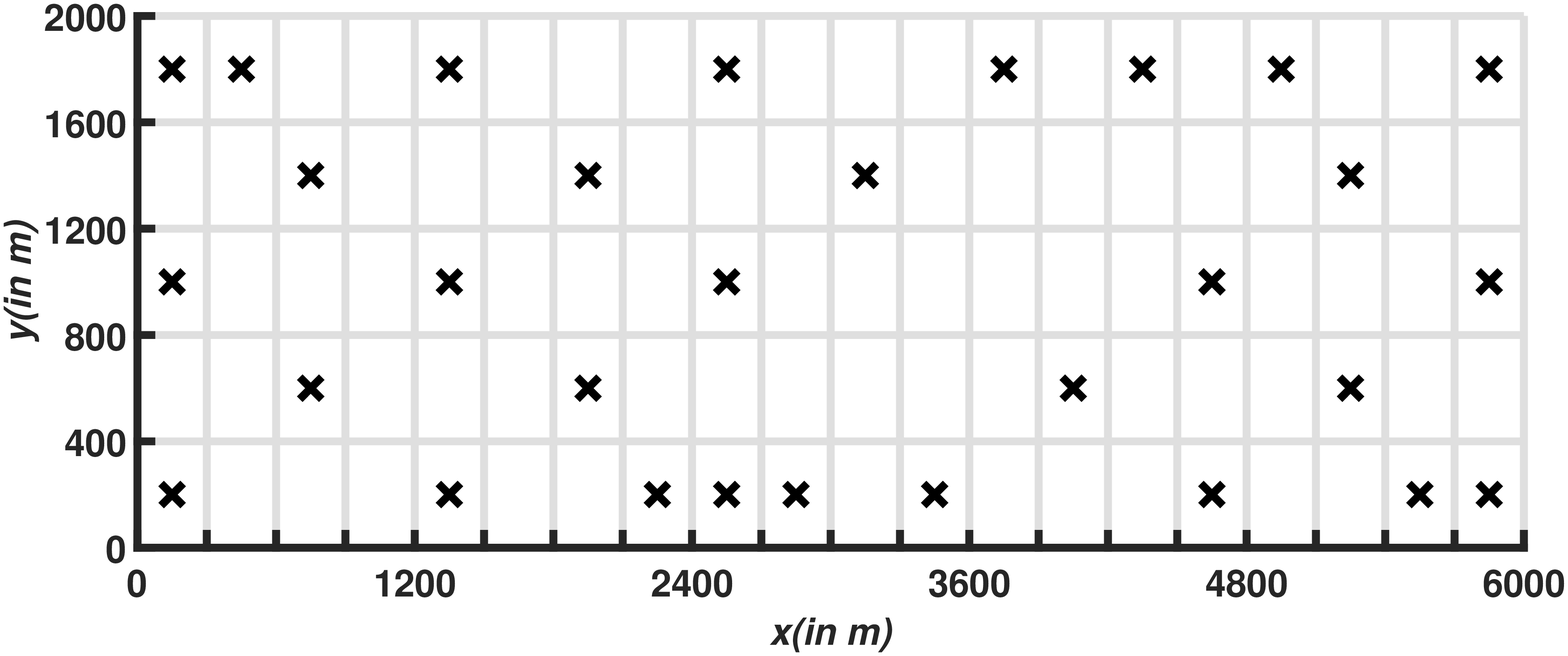}}
 \subfigure[Optimal layout for $40$ turbines]{
\includegraphics[height=3.1cm, width=11.6cm]{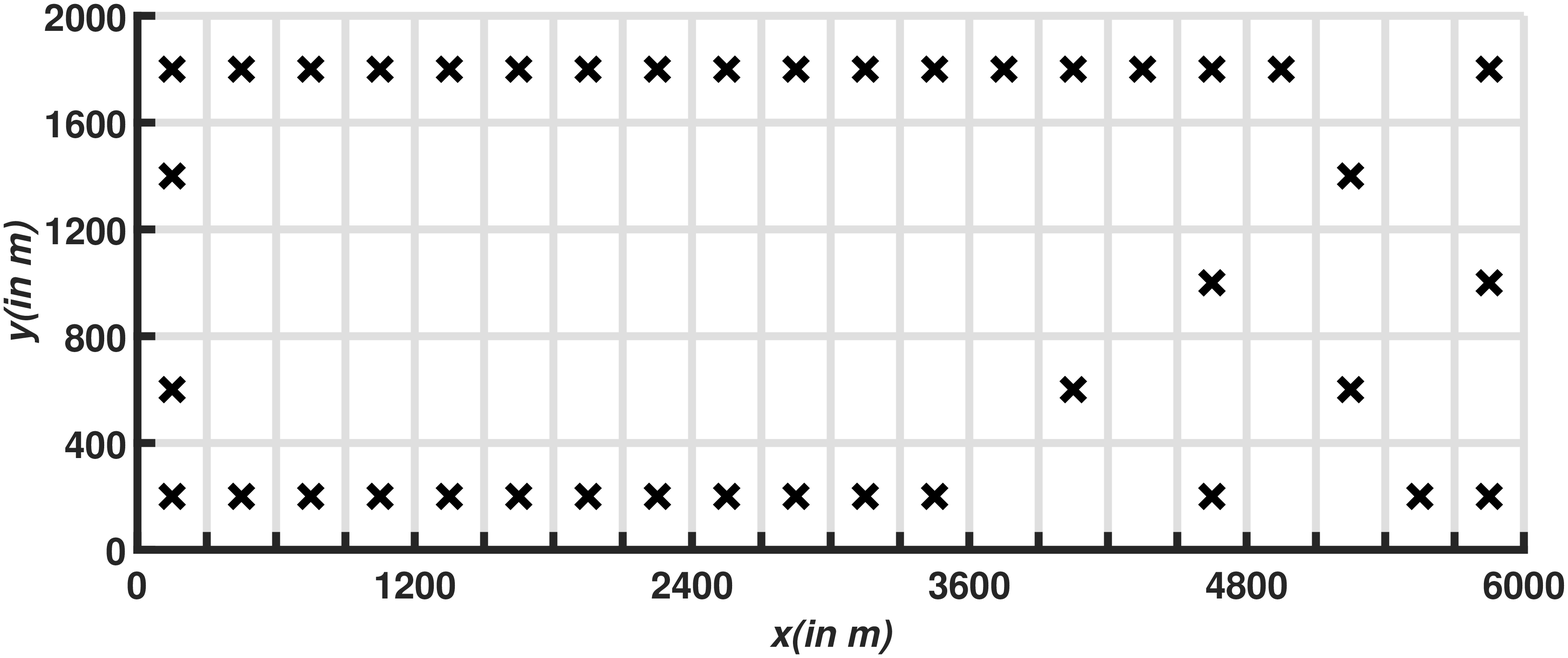}}
  \subfigure[Optimal layout for $50$ turbines]{
    \includegraphics[height=3.1cm, width=11.6cm]{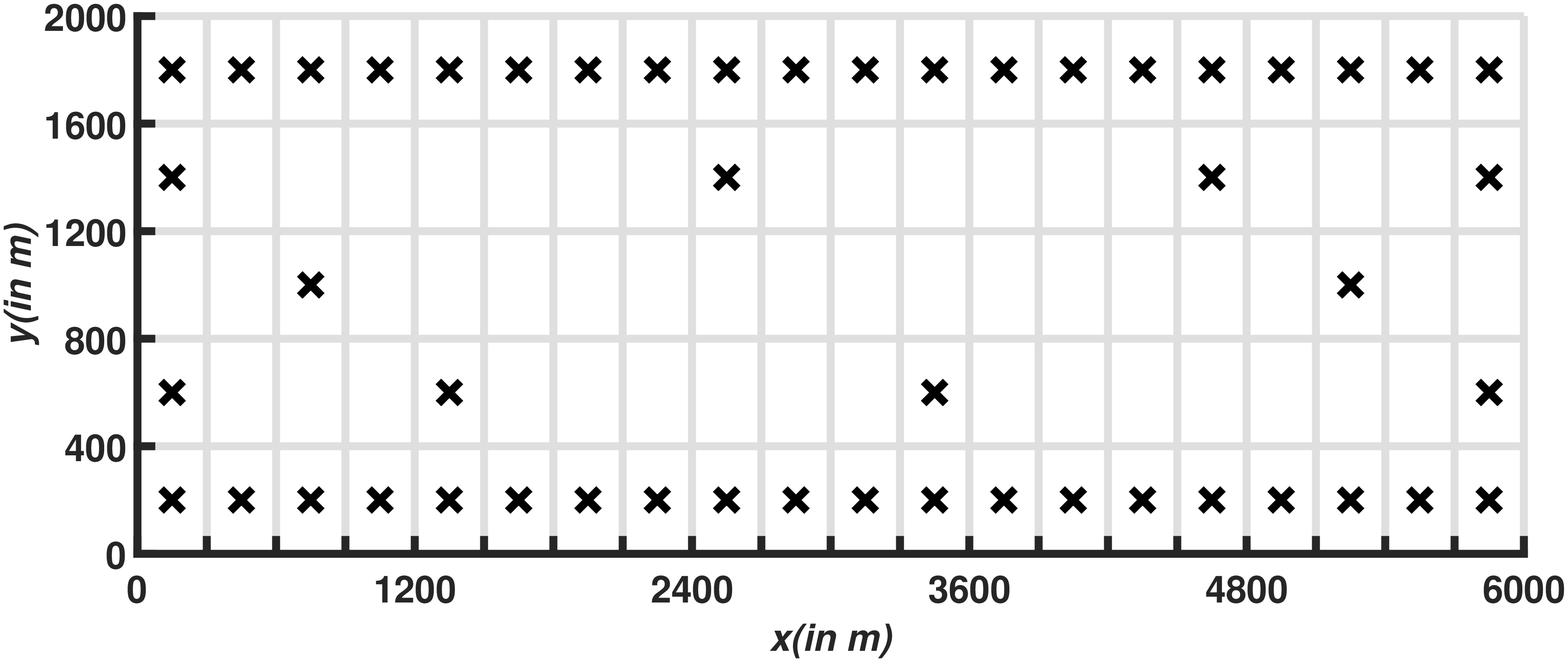}}
  \subfigure[Optimal layout for $60$ turbines]{
   \includegraphics[height=3.1cm, width=11.6cm]{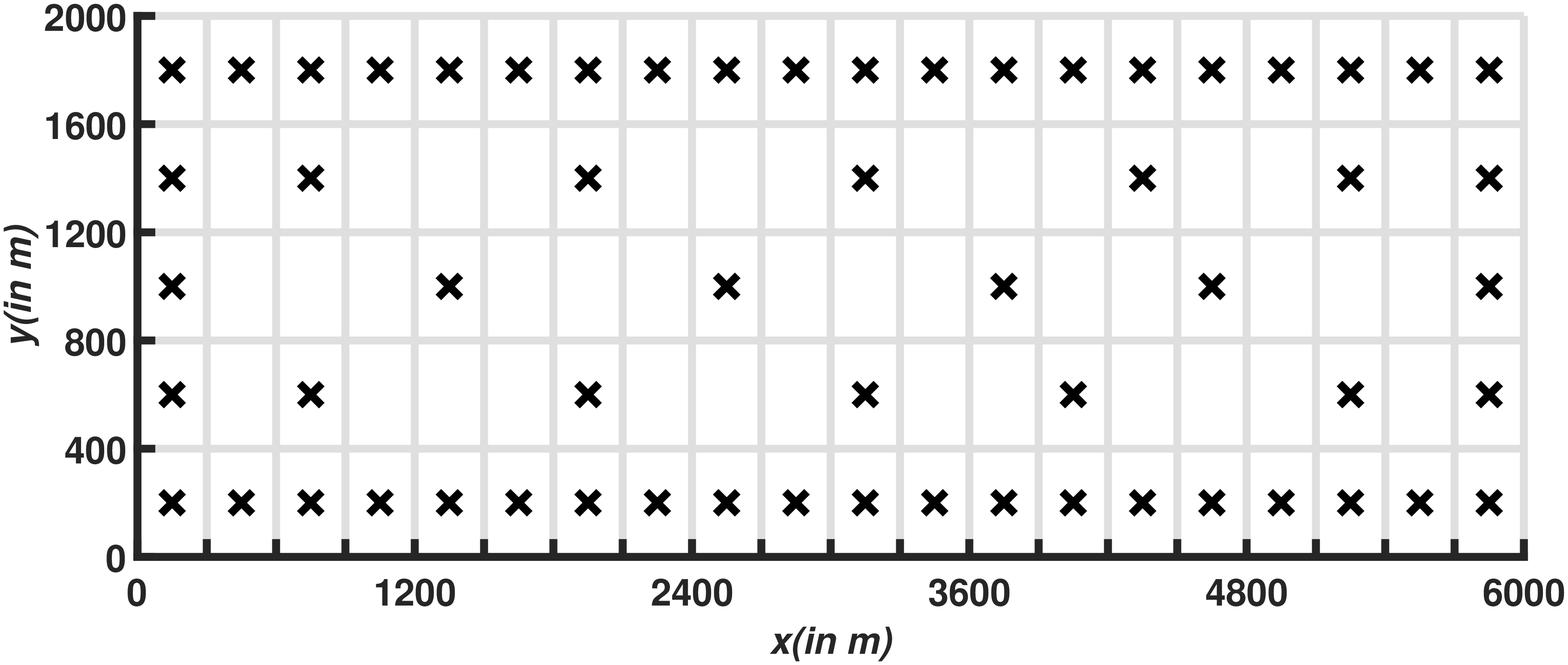}}
  \caption{Optimal layouts obtained by BNAGGSA for $N_T$ number of turbines of case 1 (site 1) } 
 \label{fig:layout1}
  \end{figure}

\begin{figure} 
\centering
  \subfigure[Optimal layout for $10$ turbines]{
    \includegraphics[height=3.1cm, width=11.6cm]{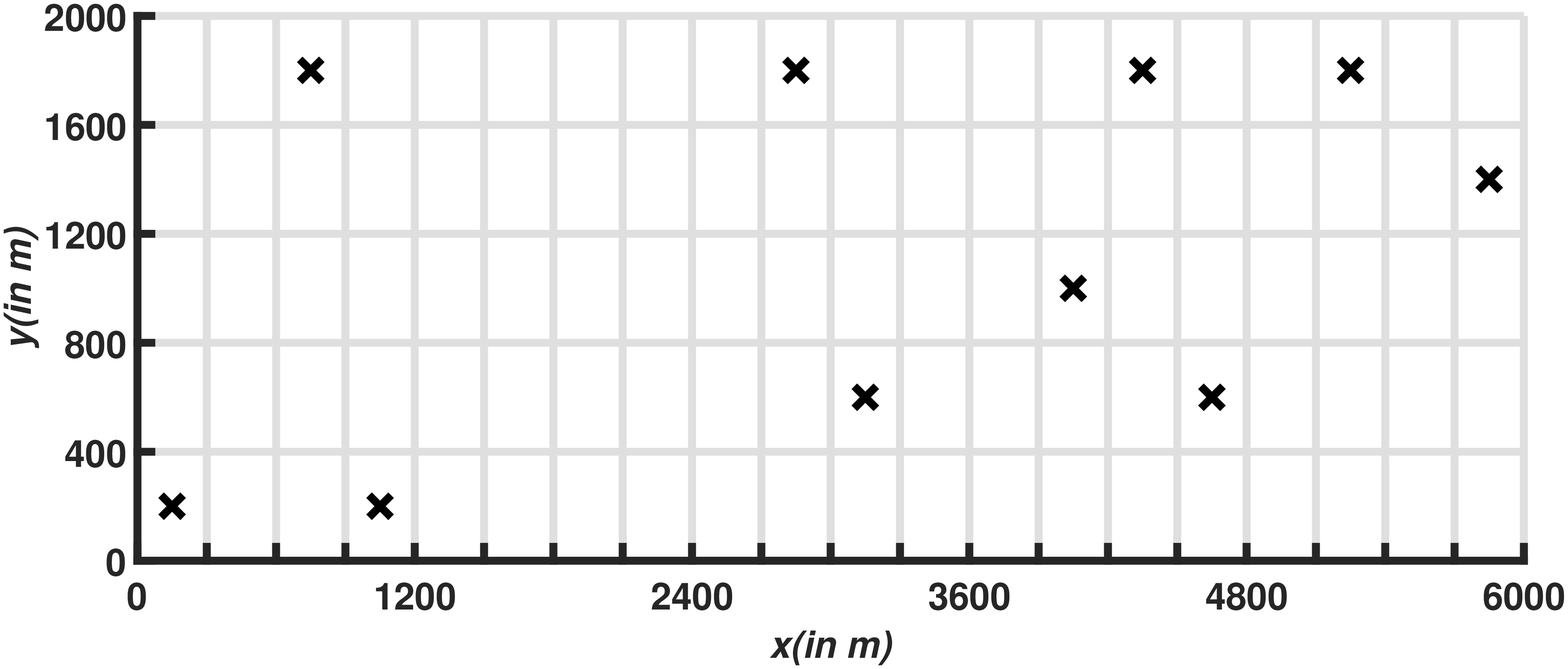}}
 \subfigure[Optimal layout for $20$ turbines]{
     \includegraphics[height=3.1cm, width=11.6cm]{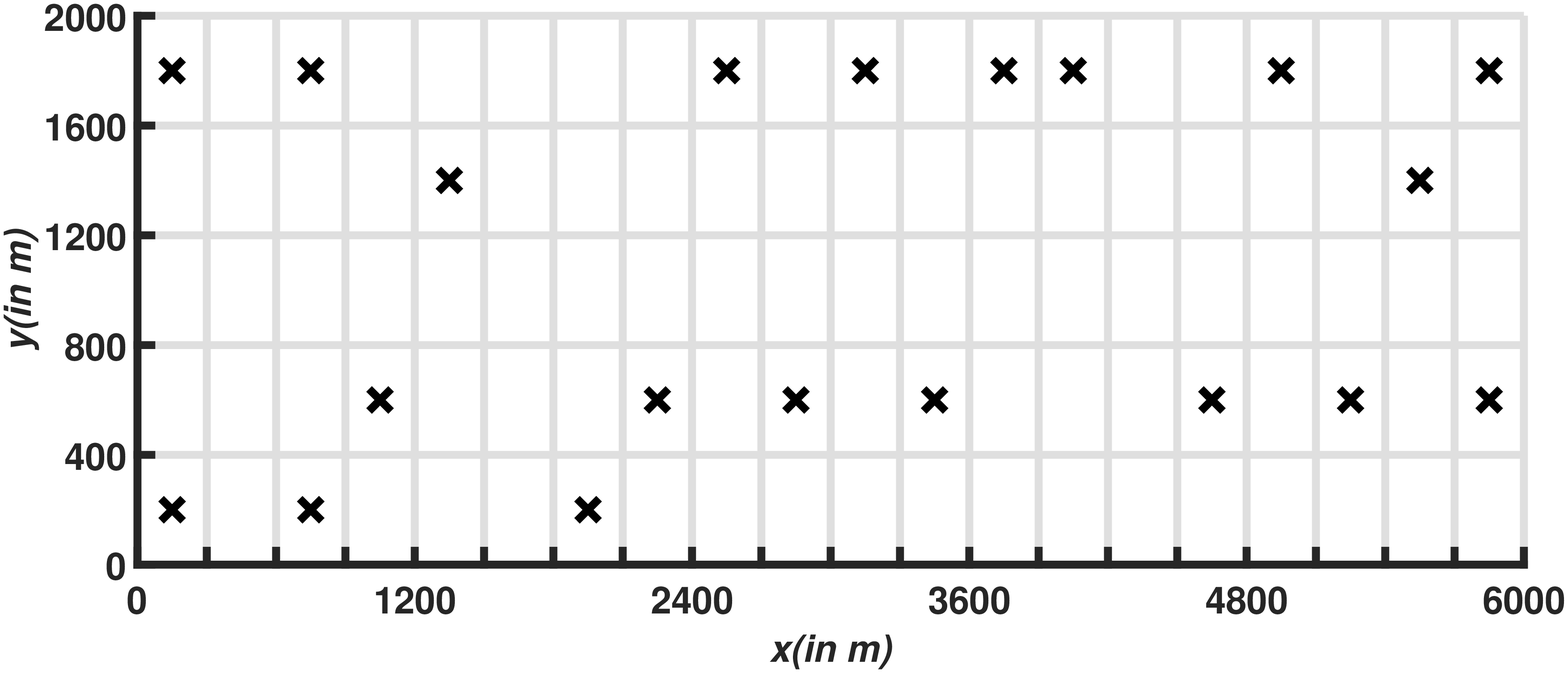}}
 \subfigure[Optimal layout for $30$ turbines]{
   \includegraphics[height=3.1cm, width=11.6cm]{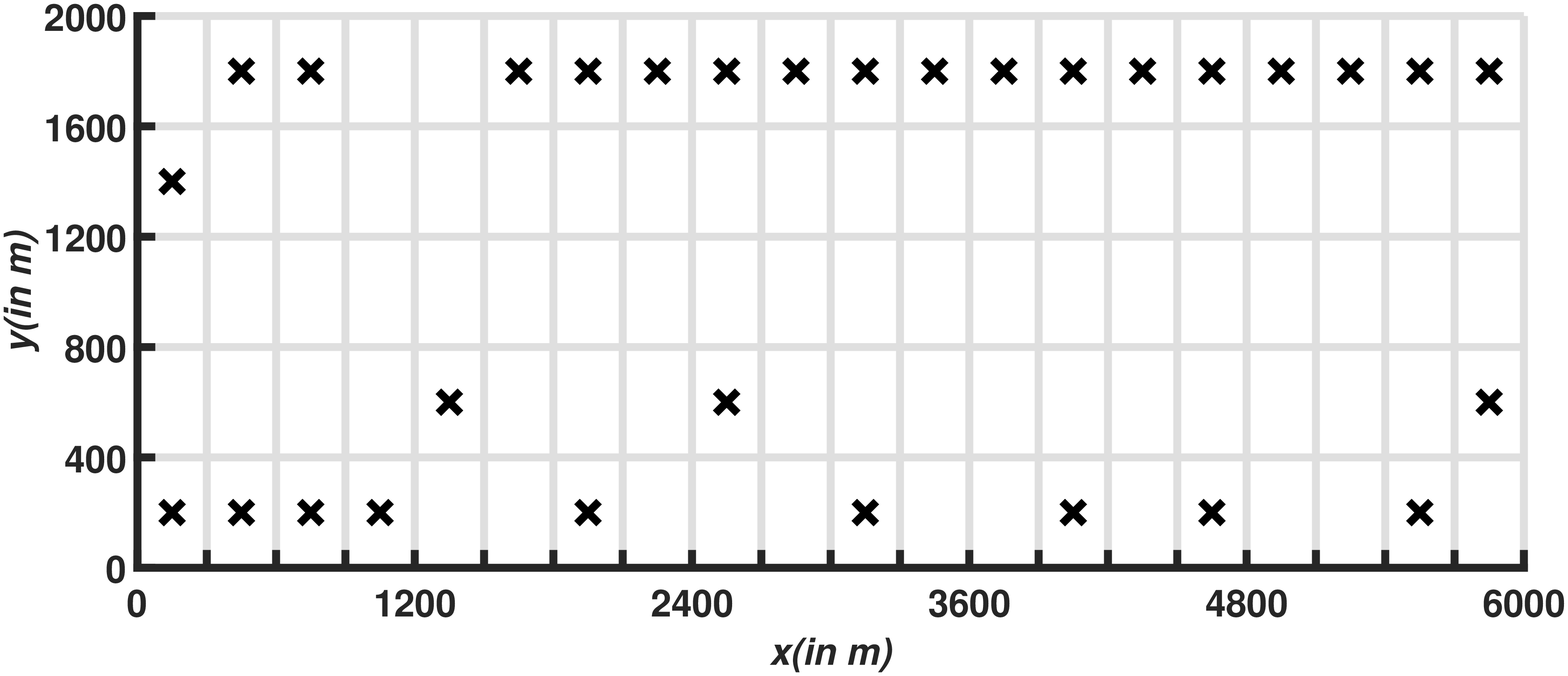}}
    \subfigure[Optimal layout for $40$ turbines]{
   \includegraphics[height=3.1cm, width=11.6cm]{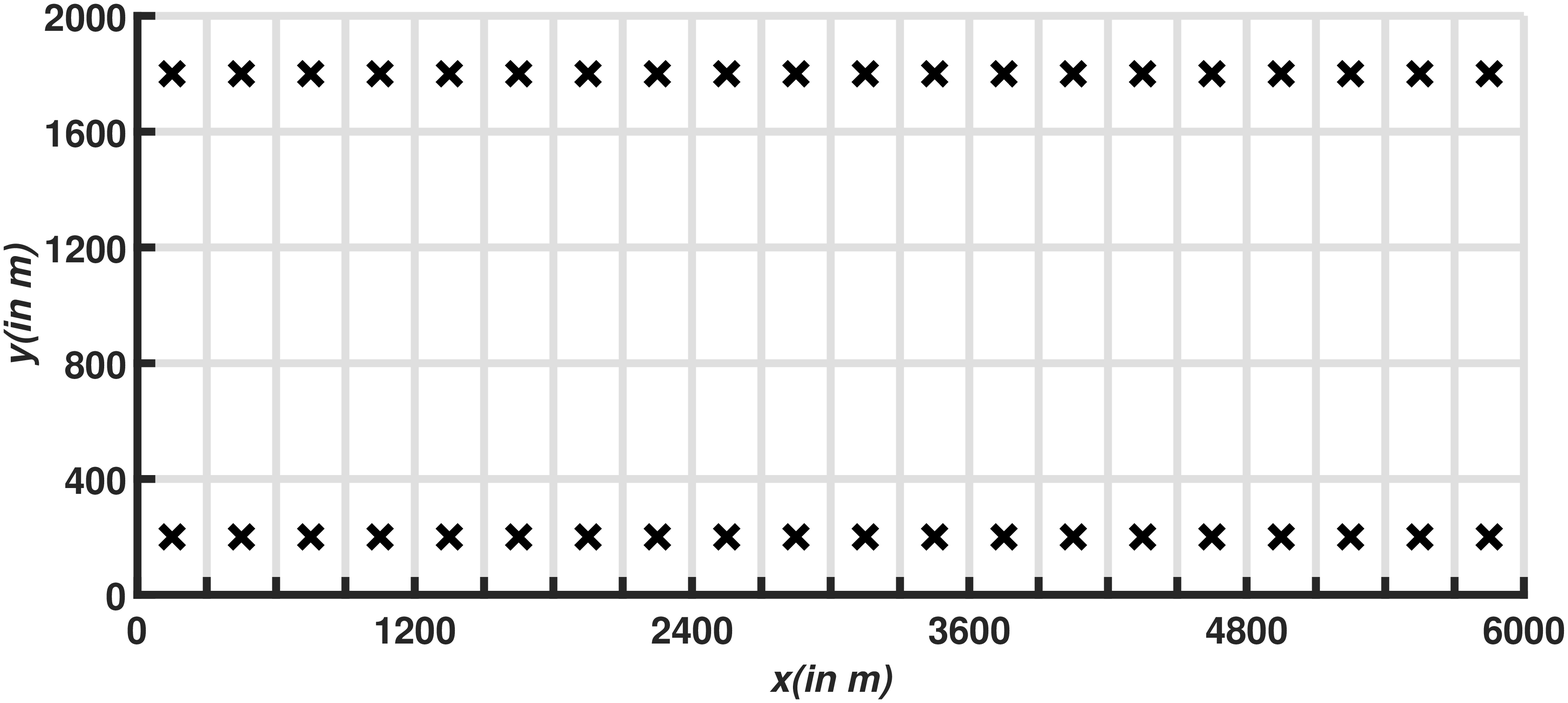}}
 \subfigure[Optimal layout for $50$ turbines]{
   \includegraphics[height=3.1cm, width=11.6cm]{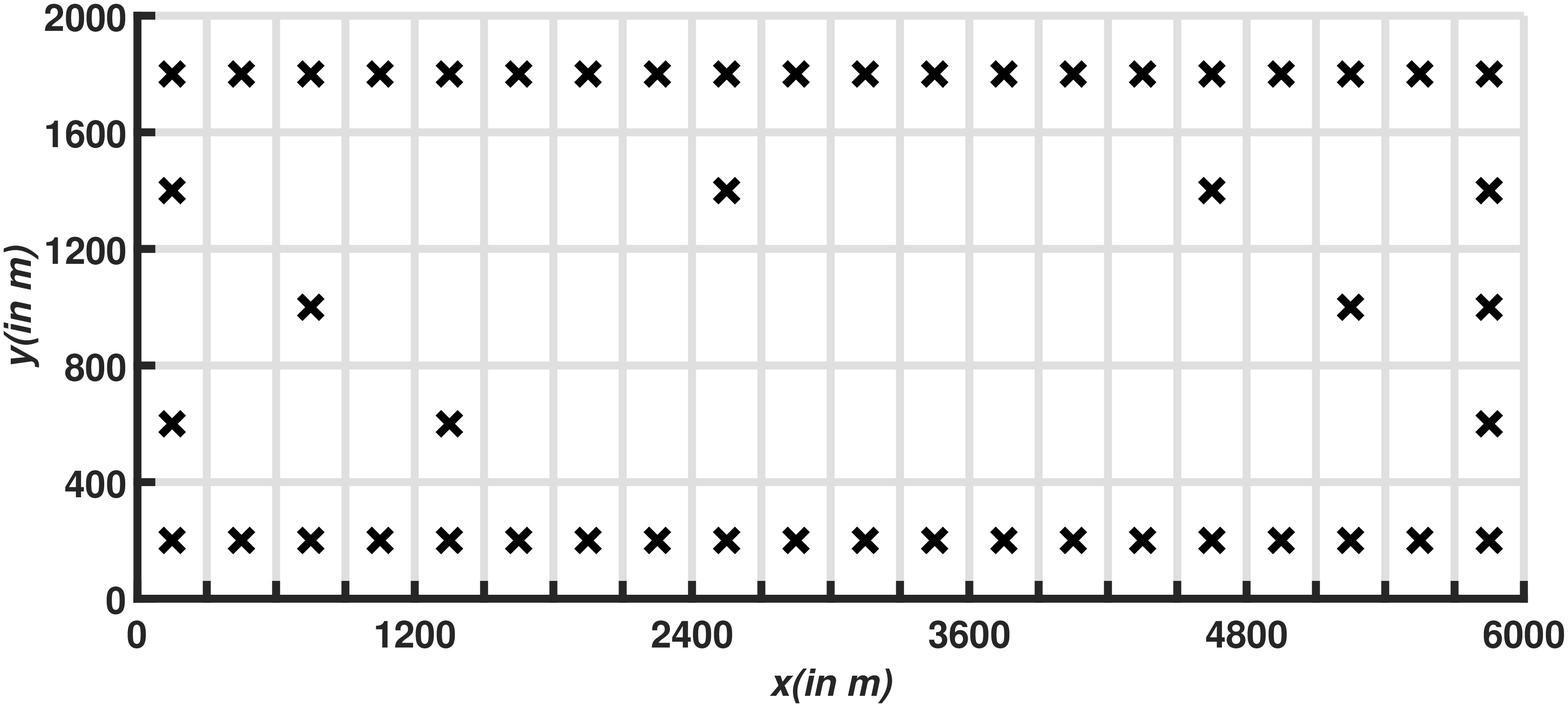}}
\subfigure[Optimal layout for $60$ turbines]{
   \includegraphics[height=3.1cm, width=11.6cm]{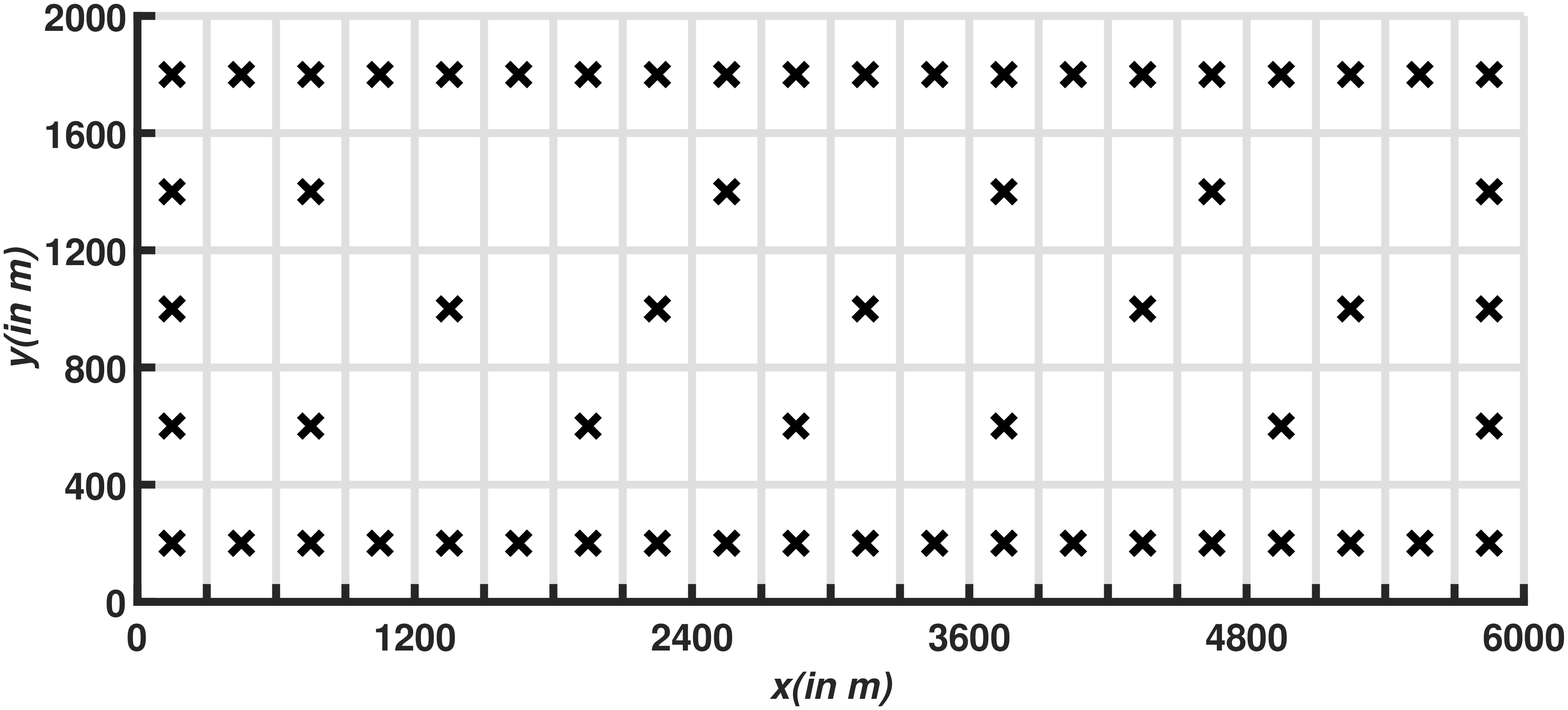}}
  \caption{Optimal layouts obtained by BNAGGSA for $N_T$ number of turbines of case 2 (site 2) } 
 \label{fig:layout2}
  \end{figure}

 Fig. \ref{fig:layout1} and Fig. \ref{fig:layout2} indicate the optimal layouts of $N_T$ turbines obtained by BNAGGSA for wind conditions at site $1$ (case 1) and site $2$ (case 2), respectively.

\begin{center}
\begin{small}
\setlength{\LTleft}{-20cm plus -1fill}
\setlength{\LTright}{\LTleft}
\begin{longtable}{p{1cm}p{3.0cm}p{2.3cm}p{2.3cm}p{2.3cm}p{2.3cm}p{2.3cm}}
 \caption{Optimized layouts results for site $1$}
 \label{table:state1}
\endfirsthead
\caption* {\textbf{Table \ref{table:state1} Continued:}}
\endhead
\toprule
\textbf{$N_T$ } & \textbf{metrics} &  \textbf{BGSA}&  \textbf{BFVGGSA}  & \textbf{BPTGSA}  &\textbf{MOEA/D}  &	\textbf{BNAGGSA} \\
 \midrule

 \multirow{3}{*} {$10$}	&	Power output (MW)	&	5.1683	&	5.1683	& 5.1683	 &     \textbf{ 5.21}           & 5.179361		\\	&	Efficiency	&	98.67\%	&  98.67\%	&	98.67\%  & \textbf{99.5\%}  &  98.89\%	\\  & Capacity factor &   25.84\%     &   25.84\%     &  25.84   & \textbf{ 26.1\%} &  25.89\%       \\[1ex]

 \multirow{3}{*} {$20$}	&	Power output (MW)	&	9.9969	&	9.9969	& 10.01179	 & \textbf{ 10.18}   &  10.1559		\\	&	Efficiency	&	95.43\%	&  95.43\%	& 95.57\%	 & \textbf{97.1\%}  &  96.94\%	\\  & Capacity factor &  24.99\%     &  24.99\%   & 25.03    & \textbf{ 25.5\%}   & 25.39\%      \\[1ex]

 \multirow{3}{*} {$30$}	&	Power output (MW)	&	14.55	&	14.55	&	14.54  & 14.78    & \textbf{14.85176}		\\	&	Efficiency	&	92.61\%	&  92.61\%	&	  92.55  & 94.1\%     &   \textbf{94.51\%}	\\  & Capacity factor &  24.25\%      &  24.25\%  & 24.24\%     &      24.6\% &  \textbf{24.75\%}       \\[1ex]

 \multirow{3}{*} {$40$}	&	Power output (MW)	&	18.88	&	18.90	& 18.87     &	 19.16    &\textbf{19.22366}		\\	&	Efficiency	&	90.09\%	&  90.21\%	&90.06\%	& 91.4\% &\textbf{91.75\%}	\\  & Capacity factor & 23.59\%       & 23.63\% &   23.59\% &  23.9\%   & \textbf{24.02\%}      \\[1ex]

 \multirow{3}{*} {$50$}	&	Power output (MW)	&	22.97	&	23.06	& 22.93	& 23.40 &\textbf{23.44308}		\\	&	Efficiency	&	87.70\%	&  88.04\%	& 87.55\% & 89.4\%  &\textbf{89.51\%}	\\  & Capacity factor &  22.97\%      &   23.06\% & 22.93\%    &  23.4\% & \textbf{23.44\%}       \\[1ex]
 
 \multirow{3}{*} {$60$}	&	Power output (MW)	&	26.96	&	26.98	& 26.90	 & 27.38   &\textbf{27.41758}		\\	&	Efficiency	&	85.79\%	&  85.85\%	&     85.62&   87.1\%	&\textbf{87.24\%}	\\  & Capacity factor &  22.46\%      &  22.49\% &22.42\%     &  22.8\% & \textbf{22.85\%}        \\[1ex]
  
\bottomrule
\end{longtable}
 \end{small}
 \end{center} 

\begin{center}
\begin{small}
\setlength{\LTleft}{-20cm plus -1fill}
\setlength{\LTright}{\LTleft}
\begin{longtable}{p{1cm}p{3.0cm}p{2.3cm}p{2.3cm}p{2.3cm}p{2.3cm}p{2.3cm}}
 \caption{Optimized layouts results for site $2$}
 \label{table:state2}
\endfirsthead
\caption* {\textbf{Table \ref{table:state2} Continued:}}
\endhead
\toprule
\textbf{$N_T$ } & \textbf{metrics} &  \textbf{BGSA}&  \textbf{BFVGGSA}  &  \textbf{BPTGSA}& \textbf{MOEA/D}  &	\textbf{BNAGGSA} \\
 \midrule

 \multirow{3}{*} {$10$}	&	Power output (MW)	&	7.926149	&	7.918980523	&	 7.932965 & \textbf{   7.98 }          & 7.949118403		\\	&	Efficiency	&	98.81\%	&  98.72\%	& 98.89\%	&\textbf{99.5\%}  &  99.09\%	\\  & Capacity factor &  39.63\%      &       39.59\% &  39.67\%   & \textbf{ 39.9\%} &  39.75\%       \\[1ex]

%
 
 \multirow{3}{*} {$20$}	&	Power output (MW)	&	15.39293	&	15.37645	&	 15.37053 & \textbf{15.61}   & 15.50229859		\\	&	Efficiency	&	95.95\%	&  95.85\%	& 95.81\%	 & \textbf{97.3\%}  &  96.63\%	\\  & Capacity factor &   38.48\%     &        38.44\% & 38.43\%  & \textbf{39.0\%}   & 38.76\%      \\[1ex]

 \multirow{3}{*} {$30$}	&	Power output (MW)	&	22.48552	&	22.50635	&	22.5028  & \textbf{ 22.97}    & 22.96395358		\\	&	Efficiency	&	93.44\%	&  93.53\%	&	93.51\%  & 95.4\%     &   \textbf{95.43\%}	\\  & Capacity factor & 37.48\%      &   37.51\%     & 37.50\%  & \textbf{ 38.3\%} &  38.27\%       \\[1ex]
 
%
%

 \multirow{3}{*} {$40$}	&	Power output (MW)	&	29.16514	&	29.35264	&	29.19196 & 30.09    &\textbf{30.11622022}		\\	&	Efficiency	&	90.89\%	&  91.48\%	& 90.98\%	 & 93.8\% &\textbf{93.86\%}	\\  & Capacity factor &    36.46\%    &  36.69\% & 36.49\%     &  37.6\%   & \textbf{37.65\%}      \\[1ex]

 \multirow{3}{*} {$50$}	&	Power output (MW)	&	35.59742	&	35.65921	&	 35.5257  &  36.50  &\textbf{36.51936845}		\\	&	Efficiency	&	88.76\%	&  88.90\%	& 88.58\% &  91.0\%  &\textbf{91.05\%}	\\  & Capacity factor &  35.59\%      &  35.66\% & 35.53\%    &  36.5\% & \textbf{36.52\%}       \\[1ex]
 
 \multirow{3}{*} {$60$}	&	Power output (MW)	&	41.61174	&	41.83223	&	41.59735 & 42.47   &\textbf{42.49998996}		\\	&	Efficiency	&	86.46\%	&  86.92\%	& 86.43\% & 88.2\%	&\textbf{88.30\%}	\\  & Capacity factor &  34.68\%      &   34.86\% & 34.67\%    &  35.4\% & \textbf{35.42\%}        \\[1ex]
  
\bottomrule
\end{longtable}
 \end{small}
 \end{center}

Therefore, the proposed BNAGGSA proves its applicability from all the considered GSA variants in the binary search space. Additionally, Despite using the single objective framework, the proposed BNAGGSA proves its supremacy over multi objective framework based MOEA/D for $30$, $40$, $50$ and $60$ turbines layout optimization of the considered windfarm in both mentioned wind sites.

\section{Conclusion}\label{sec:con}

This paper proposes a novel binary GSA variant named `A novel neighbourhood archives embedded gravitational constant in GSA for binary search space (BNAGGSA)'. In BNAGGSA, first two social interaction schemes are proposed for more diversified search mechanism
in GSA under binary search space. Then, a novel fitness-distance ratio based gravitational constant is proposed to provide a self adaptive scaling to each particle as per its current search requirements. The performance of the BNAGGSA is compared with the two variants of BGSA on $23$ popular benchmark test problems. The results show that BNAGGSA outperformed BGSA's variants in finding optimal solutions with better convergence speed. Additionally, a windfarm layout optimization problem with two different wind data sets (two case studies) has been solved by BNAGGSA. BNAGGSA obtained the most efficient layouts for any selected number of turbines within a specified range. Numerical experiments conclude that the BNAGGSA is able to find the better optimal placement of wind turbines (for more than $30$ turbines) for both case studies compare than some recent binary variants of GSA as well as MOEA/D. Thus, the proposed BNAGGSA is recommended as an efficient solver for windfarm layout optimization problem. 

\section*{Declaration}
 \textbf{Conflicts of interest/Competing interests:} All the authors declare that they has no conflict of interest.\\
  \textbf{Availability of data and material:} Not applicable.\\
  \textbf{Code availability:} Not applicable.

\end{document}